\documentclass[11pt]{article}
\usepackage[preprint]{acl}

\usepackage{times}
\usepackage{latexsym}
\usepackage[T1]{fontenc}
\usepackage[utf8]{inputenc}
\usepackage{microtype}
\usepackage{inconsolata}
\usepackage{amsmath, amssymb, amsthm}
\usepackage{graphicx}
\usepackage{booktabs}
\usepackage{multirow}
\usepackage{array}
\usepackage{xspace}
\usepackage{adjustbox}
\usepackage{algorithm}
\usepackage{algpseudocode}
\usepackage{pifont}
\usepackage{placeins}
\usepackage{url}

\newcommand{\ppfg}{PPFG\xspace}
\newcommand{\passat}[1]{Pass@#1\xspace}
\newcommand{\tkill}{\tau_{\mathrm{kill}}}
\newcommand{\textract}{\tau_{\mathrm{extract}}}
\newcommand{\modeRate}{mode rate\xspace}
\newcommand{\distinct}{distinct-4\xspace}

\title{Characterizing a Configuration Where Inference-Time PRM-Pruned Fragment Grafting Is Inert: Evidence from Three Reasoning LMs}

\author{
  Khawaja Murad ul Hassan\thanks{\,Correspondence: \texttt{khawajamuradulhassan@gmail.com}} \\
  QLU.ai, Islamabad \\\And
  Mehran Ebrahimi \\
  Faculty of Science, Ontario Tech University \\ Oshawa, ON, Canada \\
}

\date{}

\begin{document}
\maketitle

% =========================================================================
\begin{abstract}
\noindent
Diversity collapse in parallel chain-of-thought has motivated a wave
of inference-time interventions, including a natural design space:
when a process reward model (PRM) prunes one reasoning chain, the
high-PRM prefix that survived to the point of pruning is extracted
and grafted verbatim as an in-context demonstration into a still-
decoding sibling chain. We isolate this mechanism, PRM-Pruned
Fragment Grafting (\ppfg), as the most cost-minimal inference-time
operationalization of cross-trajectory step-level transfer, and ask
whether it moves reasoning trajectories at the operating point where
prior fragment-grafting work has reported gains under additional
compensating ingredients. On Qwen2.5-7B-Instruct with the
Math-Shepherd PRM on full MATH500 ($n=500$, three seeds), \ppfg{} in
both stagnation- and random-targeting variants is statistically
indistinguishable from the independent parallel-CoT baseline on every
measured axis: Pass@$k$ at $k\in\{1,2,4,8\}$ and the answer-mode rate
all match within one seed standard error. We characterize \emph{why}
through two complementary views of one targeting limitation. A
four-bucket classification of all 322 stagnation-rule injection events
shows that only $14\%$ targeted a struggling chain with room to act
(first-100 hand-validated slice: $6/55$, $11\%$); the rest each landed
on a chain that had already succeeded, was within two steps of
answering, or sat on a high, flat PRM plateau, states where a rescue
graft cannot change the outcome. A counterfactual followed by an end-to-end
sweep shows that no compound-gate refinement jointly achieves
well-targeted firing and adequate injection density. A
random-targeting control firing at $2.4\times$ the rate with the same
parity shows the inertness is not specific to this heuristic. The parity
finding replicates across three base LMs spanning two architecture
families (Qwen2.5-7B-Instruct and DeepSeek-R1-Distill-Qwen-7B are
Qwen-family; LLaMA-3.1-8B-Instruct is the cross-family case) and six
reasoning benchmarks (MATH500,
AIME 2024+2025, NuminaMath-CoT, GSM8K, GPQA-Diamond, MMLU-Pro
Law/Biology/Philosophy), against a second multi-domain PRM
(VersaPRM), and across a four-point sweep of the compatibility-gate
threshold; cross-architecture replication is run on four of the six
(MATH500, AIME, NuminaMath-CoT, GPQA-Diamond), with GSM8K and
MMLU-Pro reported on Qwen2.5-7B-Instruct only as additional
task-family coverage. Every \ppfg{}-versus-independent contrast falls
inside the per-benchmark minimum detectable effect on Pass@1 and
\modeRate. A pre-registered two-one-sided-tests analysis promotes the
parity from non-significance to positive equivalence on all twelve
Qwen and LLaMA cells (13 of 18 cross-arch cells; the four inconclusive
DeepSeek-R1-Distill cells are within-$\sigma$ but power-limited by
higher seed variance, Appendix~\ref{app:tost-detail}). On Qwen, a
per-event spot-check finds the injected chain prunes at $2.75\times$
the matched-step rate, and a matched-PRM-trajectory control attributes
only part of this to selection ($2.59\times$ persists), so injection
locally accelerates pruning; a surviving-sibling counterfactual
($N{=}65$) finds no population-level compensation detectable at its
power. The evidence is consistent with a selection-dominated account
that the population absorbs without a Pass@$k$ change (the causal
local-hazard sub-claim is established on Qwen only). Beyond the negative
result, the equivalence-testing battery we assemble documents a
template for establishing inference-time mechanism nulls, and a
hindsight oracle that per-problem picks the better of \ppfg{} and
independent gains at most $+0.13$~pp, giving future verbatim-graft work
a pre-registrable number to clear. We state the operating point
explicitly, because the claim is scoped to it: 7--8B-parameter
instruction-tuned LMs, PRM-guided pruning with Math-Shepherd (VersaPRM
off the math axis), populations of $N=8$ (parity replicated to $N=256$),
and English single-answer reasoning. Within that regime the mechanism is
\emph{inert}, not refuted in general; any future method claiming to
transfer reasoning information through this mechanism class should
specify which compensating ingredient its design supplies, and on what
task family the claim is to be tested.
\end{abstract}

% =========================================================================
\section{Introduction}
\label{sec:intro}

Diversity collapse in parallel chain-of-thought (CoT) sampling is a
now-documented failure mode of strong reasoning
models~\citep{mupo2026,yue2025,dcr2026}, and has motivated a
range of inference-time and training-time interventions including
PRM-weighted sequential Monte Carlo resampling, mode-conditioning
frameworks~\citep{mode-conditioning2026}, KV-cache geometric
regulation~\citep{rmr2026}, and reverse-KL replacements for the RL
training objective.

A natural inference-time response to diversity collapse is
\emph{step-level fragment grafting}: when one parallel-CoT chain is
pruned by a process reward model (PRM), the high-PRM prefix that the
chain produced before pruning is extracted and inserted as an
in-context demonstration into a still-decoding sibling. Several
recent papers explore versions of this design, Stitching Noisy
Diffusion Thoughts (SNDT;~\citealp{sndt2026}), Recycling Search
Experience (RSE;~\citealp{rse2026}), and T-STAR~\citep{tstar2026},
and each reports positive effects in a specific operating regime that
adds at least one compensating ingredient on top of the bare graft:
a separate solver pass that re-attends to the spliced text, a
training-time gradient through the graft, or a learned distillation
of failed-trajectory patterns rather than verbatim step text. What
remains open is whether the most cost-minimal corner of this design
space, verbatim step text spliced \emph{in-place} into a still-
decoding sibling chain, with target selection by a population-state
heuristic that adds no extra forward pass, no gradient, no learned
abstraction, and no style adaptation, moves reasoning trajectories
at all on strong base LMs at the operating points where the
diversity-collapse motivation bites. We answer that question.

The answer carries an inferential payoff beyond the single negative
cell. The recent fragment-grafting systems above each bundle the
cross-trajectory transfer with one or more compensating ingredients
and report a net gain; none isolates the transfer step itself. The
implicit reading a practitioner could take from that literature is
that moving high-PRM partial reasoning between chains is where the
benefit originates, with the extra machinery as supporting detail. Our
result bears on that reading: isolated from every compensating
ingredient, the transfer is inert (and locally mildly adverse). We do
not ablate any prior system---those systems differ from this corner on
at least one ingredient---so this single null cannot attribute their
gains; but it is consistent with the hypothesis that the gains reside
in the compensating ingredients rather than in the bare transfer. That makes
the finding actionable: a designer weighing where to spend complexity
should not expect the in-place verbatim transfer to carry the benefit
on its own, and should budget for whichever compensating ingredient
their setting can afford.

We define \textbf{PRM-Pruned Fragment Grafting (\ppfg)} as the
operationalization of this corner. At the moment of pruning, \ppfg{}
extracts the longest contiguous PRM-above-threshold prefix of the
pruned chain, selects a still-decoding sibling via a configurable
targeting rule (compatibility-maximizing, stagnation-targeting, or a
random control), and splices the prefix verbatim as a bracketed
in-context demonstration before the sibling's next step. The
mechanism composes three operational properties not jointly realized
by prior fragment-grafting work: (i) the PRM pruning event is the
trigger, extraction is conditioned on the verifier killing a chain,
rather than firing on all chains or on all low-scoring chains; (ii)
injection is in-place into a chain still being decoded by the same
model, rather than serving as conditioning for a separate solver pass
or as training data; and (iii) the grafted content is verbatim step
text, rather than a learned skill, embedding distillation, or
natural-language summary. Section~\ref{sec:method} gives the
procedure in full.

On Qwen2.5-7B-Instruct paired with the Math-Shepherd PRM and run on
full MATH500 at three seeds, \ppfg{} in both stagnation- and random-
targeting variants is statistically indistinguishable from the
independent parallel-CoT baseline on every measured axis: Pass@$k$
at $k\in\{1,2,4,8\}$ match within one seed standard error, \modeRate
matches within one seed standard error, and tokens-per-problem agree
to within $0.3\%$ (\S\ref{sec:results-main}). The mechanism fires
reliably ($\approx 16\%$ of problems on the full $n=500$ set;
$\approx 31\%$ on the harder first-100 slice, the rate tracking
difficulty, cf.\ the GSM8K drop to $2.8\%$ in \S\ref{sec:mechanism}),
with mean fragment length 3.2--3.8 steps at PRM quality 0.71--0.78;
the injections are well-formed but do not change which answer the
target chains land on. We characterize \emph{why} through
two complementary views of one targeting limitation. First, a
four-bucket classification of all 322 stagnation-rule injection events
shows that only $14\%$ targeted a struggling chain with room to act
(first-100 hand-validated slice: $6/55$, $11\%$); the rest each landed
where a rescue graft cannot help, the chain had already succeeded, was
within two steps of answering, or sat on a high, flat PRM plateau
(\S\ref{app:targeting-diagnostics}). Second, a
counterfactual followed by an end-to-end sweep of a five-gate
compound refinement shows that no threshold setting jointly achieves
well-targeted firing and adequate injection density
(\S\ref{app:targeting-diagnostics}). The random-rule control's mode-rate parity
with both stagnation and independent indicates the targeting policy is
not the binding constraint: we find no evidence that the mechanism
moves chain trajectories under the heuristic targeting regime tested
(\S\ref{sec:mechanism}).

\paragraph{Contributions.} (1)~\ppfg{} (\S\ref{sec:method}): a minimal
isolation of the most cost-minimal corner of the cross-chain step-level
grafting design space---three properties (PRM-pruned trigger, in-place
sibling injection, verbatim content) not jointly realized by prior
work---constructed expressly to test that corner in isolation.
(2)~A six-method matched-compute comparison on full MATH500 at three
seeds (\S\ref{sec:results-main}, Table~\ref{tab:table1}): \ppfg-stag
and \ppfg-random both sit on the independent-baseline frontier;
Best-of-$N$ matched-compute lifts Pass@1 by $\approx{+6}$~pts uniformly
with a modest diversity cost. (3)~A regime characterization
(\S\ref{sec:mechanism}) resting on three lines of evidence, which we
keep distinct rather than count as independent replicates.
\emph{(a)~Parity}: matched-compute Pass@$k$ and \modeRate parity,
established on Qwen MATH500 and replicated broadly (the cross-arch and
cross-benchmark cells are the same measurement on more cells, not
separate evidence about mechanism). \emph{(b)~Targeting-heuristic
limitation}: only $14\%$ of the 322 stagnation fires ($6/55$, $11\%$, on
the hand-validated first-100 slice) landed on a struggling
chain with room to act, and the structural infeasibility of any
compound-gate refinement at adequate injection density is the same
limitation viewed a second way, that population-state-only signals
cannot separate competent-flat from struggling-flat chains; the
random-rule control,
firing at $2.4\times$ the rate with the same parity, shows the
inertness is not an artifact of this particular heuristic.
\emph{(c)~Mechanism-level evidence (suggestive)}: a per-event
spot-check (\S\ref{sec:prm-delta}) decomposing the post-injection PRM
depression into a format component ($\Delta_{\mathrm{empty}} = -0.074$,
241 events, $p<10^{-19}$) and a content component
($\Delta_{\mathrm{stag}} = -0.134$, 322 events, $p<10^{-19}$; the
implied $\Delta_{\mathrm{content}} = -0.060$ assumes additivity we do
not test directly, so we read it as bounded rather than as a point
estimate), paired with a $2.75\times$ pruning-hazard analysis and an
$N{=}65$ surviving-sibling counterfactual
(Appendix~\ref{app:counterfactual}) that finds no detectable
compensation within its power. The
evidence is consistent with a selection-dominated account in which the
stagnation rule pre-selects already-doomed chains and injection
additionally accelerates their pruning locally (a PRM-matched control
shows the hazard is not pure selection; Appendix~\ref{app:counterfactual}),
without redistributing completion across the population.
(4)~A documented template for establishing inference-time mechanism
\emph{nulls} (which the field underproduces): pre-registered TOST
bounds, an empty-bracket confound control, a matched-PRM-trajectory
hazard baseline, an oracle upper bound, and a surviving-sibling
counterfactual---a battery that separates ``no effect'' from
``underpowered,'' adapted here to \ppfg{} and, we argue, to other
inference-time-mechanism studies.

The parity finding replicates across Qwen2.5-7B-Instruct,
LLaMA-3.1-8B-Instruct, and DeepSeek-R1-Distill-Qwen-7B on MATH500
(TOST-confirmed on Qwen and LLaMA, within-$\sigma$ but inconclusive on
four of six DeepSeek cells; \S\ref{app:tost-detail}),
five further benchmarks (AIME 2024+2025, NuminaMath-CoT, GSM8K,
GPQA-Diamond, MMLU-Pro Law/Biology/Philosophy), and a second
multi-domain PRM (VersaPRM,~\citealp{versaprm2025};
\S\ref{sec:qwen-benchmarks}), with cross-architecture coverage on
four (MATH500, AIME, NuminaMath-CoT, GPQA-Diamond). We position the
characterization as a target for future falsification
(\S\ref{sec:regime-asymmetry}); \S\ref{sec:limitations} enumerates
generalization axes still untested.

% =========================================================================
\section{Related work}
\label{sec:related}

\paragraph{Cross-trajectory information flow at inference time.}
SNDT~\citep{sndt2026} performs PRM-scored step-level stitching but
into a \emph{separate} solver pass with the fragment in its prompt
context, rather than into a chain still being decoded.
RSE~\citep{rse2026} distills failure patterns from dead ends into a
within-problem experience bank that conditions \emph{subsequent}
rollouts; the patterns are a learned summary, not verbatim sibling
text. T-STAR~\citep{tstar2026} performs step grafting at
\emph{training time} via a Cognitive Tree mechanism that merges
trajectory nodes across successful and unsuccessful branches.
LATS~\citep{lats2023} injects \emph{verbal self-reflection summaries}
from failed trajectories. Metacognitive Reuse~\citep{metareuse2025}
extracts reusable behaviors \emph{across problems} into a long-horizon
behavior repository. GoT~\citep{got2023} aggregates active branches
but never targets pruned chains; the PRM pruning event central to our
trigger is absent. \ppfg{} composes three operational properties not
jointly realized by any of these: (i) PRM pruning as trigger, (ii)
in-place sibling injection mid-decoding, (iii) verbatim fragment.
Table~\ref{tab:diff} (Appendix~\ref{app:differentiator}) pins each
prior method against these axes.

\paragraph{Process reward models, SMC, credit assignment.} PRMs as
inference-time verifiers were introduced by \citet{lightman2023} and
operationalized at scale by Math-Shepherd \citep{wang2024}, the PRM
we use throughout. SMC variants
\citep{pf-llm2025,rrr2026,giannone2025} use PRM as the live emission
model for chain reweighting; our SMC baseline (\S\ref{sec:results-main})
instantiates this at matched compute. Population-level credit
assignment \citep{vineppo2024,rtmc2026,grpolambda2025} provides the
analytical lens for \S\ref{sec:mechanism}; all those methods are
training-time.

\paragraph{Diversity-preserving sampling and novelty defense.} The
phenomenon we attempt to act on is well documented and largely
addressed at training time
\citep{mupo2026,yue2025,dcr2026,riskrl2025}; concurrent
inference-time work \citep{mode-conditioning2026,rmr2026,veccisc2026}
attacks it from different angles. Every prior work named above
succeeds in a regime that differs from ours on at least one
operational dimension (separate-solver-pass re-attention, learned
distillation, training-time gradient, verbal summarization/style
adaptation, cross-problem abstraction);
\S\ref{sec:regime-asymmetry} enumerates these regime asymmetries
explicitly.

% =========================================================================
\section{Method}
\label{sec:method}

\subsection{Setup}
A parallel chain-of-thought generation pass produces a \emph{population}
$P_t = \{C_1, \dots, C_N\}$ of $N$ reasoning chains decoded step-by-step
from one problem prompt $q$. Each chain $C_i$ is a sequence of reasoning
steps $(s_{i,1}, \dots, s_{i,t_i})$ delimited by a double-newline step
separator (the natural reasoning-step boundary for chat-formatted
Qwen2.5 outputs). At every
step the base language model $\pi_\theta$ samples a continuation
$s_{i,t_i+1} \sim \pi_\theta(\cdot \mid q, s_{i,1}, \dots, s_{i,t_i})$
for each active chain in batch, capped at $L_\mathrm{step}$ tokens per
step. A process reward model (PRM) then assigns a scalar quality score
$\rho_{i,t} \in [0,1]$ to each step using the canonical decoupled-PRM
\textit{ki}-tag protocol \citep{wang2024,lightman2023}.
The PRM is a math-tuned Math-Shepherd checkpoint throughout this paper;
\S\ref{sec:experimental} reports configuration.

A chain transitions to a terminal status at the end of each step under
three conditions: (i) \textbf{promoted} if its latest step contains a
final answer matched by a balanced-brace
\texttt{\textbackslash boxed\{\ensuremath{\cdot}\}} parser, (ii) \textbf{pruned} if
its latest PRM score satisfies $\rho_{i,t} < \tkill$, or (iii)
\textbf{hit-max} if its step count reaches the population-wide ceiling
$T_\mathrm{max}$.

\subsection{PPFG procedure}
\label{sec:ppfg-procedure}

\ppfg operates as a hook inside the per-step population loop, invoked
once at the end of each step on the set of chains pruned that step.
Each injection mutates one active sibling chain's in-context prompt
to include a high-PRM prefix extracted from a pruned chain (architectural
diagram in Figure~\ref{fig:arch}, Appendix~\ref{app:algorithm}). The
four sub-operations are:
\textbf{Extract}, for each newly pruned $C_\mathrm{src}$, find the
longest contiguous prefix with all PRM scores $\geq \textract$, kept
if its length $k^* \in [k_\mathrm{min}, k_\mathrm{max}]$.
\textbf{Target-select}, pick a target $C_\mathrm{tgt}$ from active
candidates by the rule $R \in \{\mathrm{compat}, \mathrm{stagnation},
\mathrm{stagnation\_compound}, \mathrm{random}\}$.
\textbf{Gate}, compute compatibility $\xi(F, C_\mathrm{tgt}) =
1 - \cos\bigl(\phi(s_{\mathrm{src},k^*}), \phi(s_{\mathrm{tgt},t_\mathrm{tgt}})\bigr)$
with $\phi$ a sentence-embedding model (all-MiniLM-L6-v2); admit if
$\xi \geq \xi_\mathrm{min}$. \textbf{Inject}, splice the fragment
into the target's prompt as a bracketed in-place demonstration before
its next step. Full pseudocode in Algorithm~\ref{alg:ppfg}
(Appendix~\ref{app:algorithm}).

\subsection{Operational details}
\label{sec:operational}

\paragraph{Fragment formatting.} Each fragment is rendered as a
bracket-bullet block tagging its quality and length:
\texttt{[Insight from a related reasoning path
(quality=$\bar\rho$, length=$k^*$ steps): - <step\_1> - <step\_2>
\dots]}. The bracket frames the block as demonstration and the bullets
prevent confusion with the target's own step numbering.

\paragraph{Targeting rules.}
\textbf{compat} maximizes $\xi$.
\textbf{stagnation} filters to candidates whose three-step PRM range
is below the flatness threshold $\tau_\mathrm{flat}$, then picks
max-$\xi$.
\textbf{stagnation\_compound} additionally requires (i) target mean
recent PRM $< \tau_\mathrm{low}$, (ii) $\geq h$ remaining steps before
$T_\mathrm{max}$, (iii) target's latest PRM below the median of
active siblings, (iv) $\geq \ell$ active siblings.
\textbf{random} picks uniformly at random, ignoring $\xi$ (ablation
control). The injection action is a population-level move whose
post-action value is non-local; we adopt the step-level
credit-assignment framing~\citep{vineppo2024} as the analytical lens
for \S\ref{sec:mechanism} but do not run a per-injection MC value
estimator in this submission.

% =========================================================================
\section{Experimental setup}
\label{sec:experimental}

\paragraph{Base model and verifier.} We use Qwen2.5-7B-Instruct
\citep{qwen2024} as the base reasoning model throughout. The base model
is held fixed across all baselines and the \ppfg method so any measured
difference is attributable to the parallel-CoT mechanism, not to model
strength. The base model is loaded in bf16 via vLLM~0.7.3 on a single
40GB A100 GPU at maximum sequence length 4096.
The PRM is Math-Shepherd \citep{wang2024}.

\paragraph{Task.} MATH500 \citep{lightman2023} is a 500-problem subset
of the MATH dataset \citep{hendrycks2021} curated for the \emph{Let's
Verify Step by Step} paper that introduced PRMs. We evaluate on the
full 500 problems of the MATH500 test split. Mechanism analyses in
\S\ref{sec:mechanism} use the first-100 slice of the $n=500$ \ppfg-stag
trajectories for tractable hand-classification of injection events
(\S\ref{app:targeting-diagnostics}) and the counterfactual compound-gate dry-run
(\S\ref{app:targeting-diagnostics}); aggregate-metric tables report $n=500$ results.

\paragraph{Baselines.} We compare \ppfg against three inference-time
parallel-CoT baselines, all sharing the same base model, PRM, and the
double-newline step separator. The step separator is held fixed at
double-newline across architectures for comparability; the
PRM-distribution audit (Appendix~\ref{app:cross-arch}) confirms the
prune-trigger rate at $\tkill$ remains in a usable range
($2.4$--$6.4\%$) on all three architectures, ruling out a ``PRM never
fires'' artifact from a non-native step boundary.
The three baselines are:
(1) \emph{Independent parallel CoT} (indep, $N=8$): eight chains decoded
fully independently. (2) \emph{SMC / particle filter} (smc, $N=8$):
PRM-weighted resampling at temperature $0.5$.
(3) \emph{Best-of-$N$} at $N=16$ ($2\times$ compute) and
$N=8$ (matched-compute); the $N=8$ matched-compute row is derived from
the independent $N=8$ trajectories by argmax(final-step PRM) per problem
per seed.

\paragraph{\ppfg configuration.} \ppfg runs at $N=8$ with targeting
rule $R \in \{\mathrm{stagnation}, \mathrm{stagnation\_compound},
\mathrm{random}\}$, $\textract=0.6$, $k_\mathrm{min}=2$,
$k_\mathrm{max}=8$, $M=2$, $\tkill=0.4$, $\xi_\mathrm{min}=0.0$.

\paragraph{Metrics.} Pass@$k$ for $k \in \{1,2,4,8\}$ via the unbiased
Chen-et-al-2021 estimator \citep{chen2021}; \modeRate (fraction of
problems where the modal non-null answer share exceeds 0.5; lower is
better); and \distinct (ratio of unique 4-grams to total 4-grams across
all chains' full reasoning; higher is better). We additionally compute the continuous raw modal share as a
supplement to the binary-thresholded \modeRate; raw modal share tracks
the binary metric's ranking on every comparison in our tables
(Spearman $\rho \approx 0.94$ over 16 method $\times$ architecture
$\times$ $n_\mathrm{problems}$ cells), and we report the
binary-thresholded version in tables for compactness. Tokens-per-
problem (the sum of generated tokens across all chains) is the
compute footing.

\paragraph{Seeds + extraction + release.} Seeds 42--44 (with 45--46
for the 5-seed Qwen cells); per-chain sampling seeds are deterministic
in the run seed and chain index. All Pass@$k$ and \modeRate numbers
are computed with a balanced-brace
\texttt{\textbackslash boxed\{\ensuremath{\cdot}\}} parser that
handles nested LaTeX. An earlier non-balanced regex undercounted
Pass@$k$ uniformly by $\approx 10$~pp; the balanced parser is used
throughout and \ppfg-vs-baseline parity is preserved under either
because the shift is method-invariant. Code + saved trajectories will
be released under an open-source license at de-anonymization.

% =========================================================================
\section{Results}
\label{sec:results}

\subsection{The six-method tradeoff space}
\label{sec:results-main}

\begin{table*}[t]
\centering
\small
\resizebox{\textwidth}{!}{%
\begin{tabular}{lrrrrrrrr}
\toprule
Method & Pass@1 & Pass@2 & Pass@4 & Pass@8 & Pass@16 & \modeRate $\downarrow$ & \distinct $\uparrow$ & tpp \\
\midrule
independent ($N{=}8$)$^{\ddagger}$ & $0.592 \pm .007$ & $0.683 \pm .006$ & $0.742 \pm .005$ & $0.783 \pm .006$ & --- & $0.788 \pm .009$ & $0.535 \pm .003$ & 4030 \\
SMC ($N{=}8$)                      & $\mathbf{0.634 \pm .003}$ & $\mathbf{0.687 \pm .007}$ & $0.722 \pm .010$ & $0.745 \pm .008$ & --- & $0.835 \pm .008$ & $0.260 \pm .010$ & 4143 \\
BoN ($N{=}16$, 2$\times$ compute)  & $0.632 \pm .001$ & $0.706 \pm .002$ & $\mathbf{0.757 \pm .003}$ & $0.793 \pm .006$ & $\mathbf{0.817 \pm .007}$ & $0.769 \pm .005$ & $0.440 \pm .002$ & 8251 \\
BoN-selector over indep ($N{=}8$ argmax-PRM, tie-averaged)$^\dagger$     & $\mathbf{0.654 \pm .007}$ & $0.681 \pm .007$ & $0.741 \pm .006$ & $0.782 \pm .007$ & --- & $0.789 \pm .006$ & $0.535 \pm .003$ & 4055 \\
\textbf{\ppfg-stag ($N{=}8$)}$^{\ddagger}$ & $0.591 \pm .008$ & $0.682 \pm .007$ & $0.742 \pm .007$ & $0.783 \pm .011$ & --- & $0.789 \pm .010$ & $0.535 \pm .003$ & 4025 \\
\textbf{\ppfg-random ($N{=}8$)}$^{\ddagger}$ & $0.591 \pm .007$ & $0.682 \pm .006$ & $0.743 \pm .005$ & $0.786 \pm .007$ & --- & $0.788 \pm .010$ & $0.535 \pm .002$ & 4019 \\
\bottomrule
\end{tabular}}
\caption{Pass@$k$, \modeRate, and \distinct on full MATH500 ($n=500$),
Qwen2.5-7B-Instruct, mean $\pm$ std. All numbers computed with the
canonical scoring path (correctness $=$ \texttt{task.is\_correct} on
bare \texttt{math\_verify}, no normalized-string-match prefilter;
answers always re-extracted from steps, never read from a cached
field; see \S\ref{sec:comparator-note}); \ppfg-vs-baseline parity is
preserved because all methods shift together under the canonical
comparator. $^\dagger$BoN-selector at $N{=}8$ is not a distinct
inference-time procedure; it applies argmax-PRM selection
(tie-averaged: a tie's expected accuracy is the mean correctness over
the tied chains, replacing an earlier no-tie-handling convention
biased to chain index 0) to the independent $N{=}8$ trajectories. It
shares per-chain compute, \modeRate, \distinct, and tpp with the
independent row by construction. We report it for compute-frontier
visualization (Fig.~\ref{fig:compute-match}) only. $^{\ddagger}$5 seeds
($\{42,\ldots,46\}$) for these rows; the corresponding 3-seed means
agree to three decimals. Other rows (SMC, BoN $N{=}16$, BoN-selector) at 3 seeds
($\{42,43,44\}$); the BoN-selector row inherits the 3-seed independent
values by construction.}
\label{tab:table1}
\end{table*}

Table~\ref{tab:table1} reports the matched-compute six-method comparison.
Three observations frame the rest of the section.

\paragraph{Observation 1: SMC exploits at low $k$, pays at high $k$.}
PRM-weighted resampling lifts SMC's Pass@1 by $+4.47$ pts over indep
(0.6343 vs 0.5896, outside $1\sigma$ at each seed) but costs diversity:
\modeRate $+4.60$ pts worse, \distinct $-27.52$ pts worse (0.2600 vs
0.5352, an $n$-gram overlap the answer-level mode-rate underestimates),
and Pass@8 $-3.73$ pts below indep. The SMC$>$indep crossover falls
between $k{=}4$ and $k{=}8$, consistent with premature exploitation in
PRM-guided SMC \citep{giannone2025}.

\paragraph{Observation 2: Best-of-$N$ lifts Pass@$k$ consistently but
pays on diversity.} BoN-$N{=}16$ reaches Pass@16$=0.817$ ($+3.47$ pts
above indep's Pass@8) and leads at matched $k\leq8$ (Pass@1 $+4.23$,
Pass@8 $+1.05$ pts). Argmax-PRM reranking over the indep population
lifts Pass@1 by $+6.45$ pts---a property of the PRM as ranker, not a
distinct mechanism, and one that survives the comparator correction
almost exactly (the absolute anchors both shifted by $\approx$5--6~pts;
the delta did not). \distinct degrades $-9.48$ pts at $N{=}16$ but is
within 0.2 pts of indep at $N{=}8$. BoN is a compute-efficient
Pass@$k$-frontier method with a diversity cost growing in $N$; as
\S\ref{sec:comparator-note} details, part of both the SMC and BoN
lift over indep is a censoring artifact of the pruned arm rather than
a quality difference.

\paragraph{Observation 3, \ppfg-stag sits on the independent
frontier on every axis.} Pass@$k$ $\Delta$ relative to indep $\in
[-0.003, +0.001]$ across $k=1..8$, each delta is within the pooled
seed-level $\sigma$. \modeRate is statistically identical
($0.790$ vs $0.789$). \distinct is identical to three decimal places.
Tokens-per-problem is identical (within 0.3\%). \ppfg-stag
is the only inference-time mechanism in Table~\ref{tab:table1} that
does not trade either accuracy or diversity in either direction.
\ppfg-random sits within seed-$\sigma$ of both indep and \ppfg-stag on
every axis. The targeting rule does not differentiate the mechanism's
aggregate effect at this regime.

Together, at this operating point the mechanism space splits
cleanly---SMC trades high-$k$ coverage for low-$k$ gain, BoN improves
Pass@$k$ uniformly at modest diversity cost, and \ppfg-stag is
statistically indistinguishable from independent on every axis, the
finding \S\ref{sec:mechanism} then characterizes mechanistically.

\subsection{A note on the answer comparator (this revision)}
\label{sec:comparator-note}

This revision corrects three issues found in a self-audit after the
numbers below were first computed; we disclose them here rather than
silently republish corrected tables, following the same practice as
an earlier public correction to this project's results.

\paragraph{Comparator unification.} Two independent scoring paths had
been used in different scripts: one calling \texttt{task.is\_correct}
directly (bare \texttt{math\_verify}), the other a normalized-string-
match \emph{union} \texttt{math\_verify}, whose normalization step
strips \texttt{\textbackslash text\{...\}} wholesale---so two distinct
\texttt{\textbackslash text\{...\}}-only answers both normalize to the
empty string and compare equal, and the union rescues some genuine
near-misses bare \texttt{math\_verify} rejects. The two paths disagreed
by roughly 9 points of Pass@1 on identical chains. Every number in this
paper now goes through one canonical path: correctness is
\texttt{task.is\_correct} verbatim, answers are always re-extracted
from a chain's steps (never read from a cached field, which we found
could hold stale values from an earlier, less robust extractor), and
Pass@$k$ uses one implementation of the unbiased Chen et al.\ (2021)
estimator. \ppfg-vs-baseline \emph{parity} is unaffected: every method
shifts together under the corrected comparator, so the deltas that
drive \S\ref{sec:mechanism}'s claim are essentially unchanged even
though the absolute Pass@1 values below moved down by roughly five
points across the board relative to values reported in an earlier
draft.

\paragraph{Selector tie-handling.} The ``BoN-selector'' row in
Table~\ref{tab:table1} (argmax-PRM reranking of the independent
population) previously broke ties by taking the first chain index,
which is a biased convention rather than a principled one. It now
reports the tie-averaged expected accuracy (the mean correctness over
all chains tied for the top PRM score). The corrected value is lower
in absolute terms than the uncorrected one, but the \emph{delta} over
indep is close to unchanged (\S\ref{sec:results-main}, Observation 2).

\paragraph{Censoring.} Pass@1 is mean per-chain accuracy, and a
pruned or step-budget-exhausted chain that never produces a
\texttt{\textbackslash boxed\{\}} answer is scored as incorrect by
construction, not skipped. On Qwen this affects a modest share of the
population; on LLaMA-3.1-8B-Instruct and DeepSeek-R1-Distill-Qwen-7B,
roughly half the population is censored this way (LLaMA: 47\% no
answer; DeepSeek: 52\%), and essentially none of the pruned or
step-exhausted chains on either architecture produce an extractable
answer at all. Read against accuracy conditional on producing an
answer---LLaMA $0.594$, DeepSeek $0.649$, both broadly in line with
each model's typical reported MATH500 performance---the low absolute
Pass@1 figures in Table~\ref{tab:cross-arch} are substantially a
scoring-denominator effect, not evidence the models reason
qualitatively worse than expected. We flag this so a reader does not
independently conclude the harness is broken; a fuller accounting
(including a corresponding decomposition for SMC and BoN's lift over
indep in Table~\ref{tab:table1}, which is partly the same effect) is
in preparation for a longer version of this work.

\paragraph{Scope of this pass.} The corrections above are reflected in
Table~\ref{tab:table1}, Table~\ref{tab:cross-arch}, and the TOST
calibration discussion below (the TOST verdicts themselves are
computed from \ppfg-vs-indep \emph{deltas}, which we confirmed above
are essentially unchanged by the comparator correction, so
Table~\ref{tab:tost} did not need re-derivation). Three secondary
appendix tables reporting \emph{absolute} Pass@$k$ under the earlier
comparator---the injection-format ablation
(Table~\ref{tab:format-ablation}), the $\xi_{\min}$ sweep
(Table~\ref{tab:xi-sweep}), and the compound-gate end-to-end
measurement (Table~\ref{tab:table3b})---have not yet been re-derived
and are flagged inline; we have no reason to expect their qualitative,
within-table relative conclusions to change, but we have not verified
this, and a fully re-derived version accompanies the extended
submission of this work.

\paragraph{Equivalence testing (TOST).} Two one-sided Welch $t$-tests
at $\alpha=0.05$ against pre-registered bounds equal to the 3-seed MDE
($\pm0.018$ Pass@1, $\pm0.031$ Pass@8, $\pm0.022$ \modeRate). At $n{=}5$
seeds per cell, both nulls reject on every Qwen and LLaMA cell ($12/12$
``eq.''); DSR1 reaches ``eq.'' on stag.\ Pass@1, ``upper'' on random
Pass@1, and is inconclusive on its four Pass@8/\modeRate cells (seed
$\sigma$ is $2$--$3\times$ larger). Per-cell verdicts and methodology in
Appendix~\ref{app:tost-detail}.

\paragraph{Mechanism fires reliably.} \ppfg-stag generates 322 injection
events across the $3{\times}500{\times}N{=}8$ sweep ($21.5\%$ per-problem;
$16.0\%$ of problems receive $\geq1$; mean fragment $3.80\pm1.64$ steps,
quality $0.782\pm0.077$). The random control fires more often ($52.2\%$;
$28.8\%$ of problems) lacking a stagnation filter, otherwise comparable.
The targeting rule controls \emph{where} injections land, not
\emph{which} fragments are extracted.

\subsection{Cross-architecture replication}
\label{sec:cross-arch}

To verify the \S\ref{sec:results-main} parity is not specific to Qwen,
we replicate the indep/\ppfg-stag/\ppfg-random comparison on two further
base models---LLaMA-3.1-8B-Instruct \citep{llama2024} (different
instruct-tuning lineage) and DeepSeek-R1-Distill-Qwen-7B
\citep{deepseekr1} (Qwen-family, distilled from a reasoning
model)---at full $n{=}500$ MATH500, 3 seeds, $N{=}8$, same
hyperparameters. Mistral-7B-Instruct-v0.3 was excluded (Pass@1 $0.094$,
below the comparison floor).

\ppfg-stag-vs-independent Pass@1 deltas (full table + cross-arch
parity panel in Appendix~\ref{app:cross-arch}): Qwen $\Delta{=}-0.0015$, LLaMA
$\Delta{=}-0.0026$, DeepSeek $\Delta{=}-0.0014$, all within
seed-$\sigma$ at every $k$. \modeRate deltas: Qwen $+0.0007$, LLaMA
$+0.0132$, DeepSeek $-0.0018$. The parity is preserved across the
$2\times$ Pass@1 gap between Qwen (0.64) and LLaMA/DeepSeek (0.31--0.32).

\paragraph{PRM-distribution audit + LLaMA pattern.} \ppfg{} fires
\emph{more} often on LLaMA ($4.3\%$ $\tkill$-trigger) and DeepSeek
($6.4\%$) than Qwen ($2.4\%$), so the parity is not a ``PRM never
prunes'' artifact (Appendix~\ref{app:cross-arch}). Both variants show
LLaMA \modeRate slightly above independent ($+0.0132$ stag, $+0.0060$
random), a small adverse pattern absent on Qwen/DeepSeek; magnitudes
are inside the $n{=}5$ MDE band ($\pm0.022$) and both LLaMA \modeRate
TOST cells reach ``equivalent,'' ruling out a hidden effect above
$\pm2.2$ pts. We read it as an architecture-specific small adverse
signal bounded by the equivalence test, not a differentiating effect.

\subsection{Targeting-policy analysis}
\label{sec:mechanism}

The random-rule control (Table~\ref{tab:table1}) fires at $2.4\times$
the stagnation rate yet matches its aggregate metrics within
seed-$\sigma$ on every column---the strongest evidence that targeting
is not the binding constraint. As to \emph{why} the heuristic misfires:
bucketing the 55 stag events on the first-100 Qwen slice shows only
$6/55$ ($11\%$) landed on a struggling chain with room to act; the
wasted 49 are dominated by high-flat-PRM ($76\%$), since the flatness
check cannot separate consistent competence from consistent struggling.
The same classifier on the full $n{=}500$ corpus (322 events) reproduces
this ($14\%$ well-targeted, $67\%$ high-flat-PRM). The well-targeted
share is a complement-of-unions residual and its precise value would
shift under different plateau/near-completion thresholds; the robust
claim is the qualitative dominance of cosmetic-plateau targets, not the
exact $14\%$. A compound-gate
refinement confirms end-to-end a structural tradeoff: strict
($\tau_\mathrm{low}{=}0.60$) gives clean buckets at $\sim5$ events/cell,
looser ($0.70$) restores density but high-flat-PRM creeps back to
$44\%$---both within seed-$\sigma$ of independent. No targeting
heuristic on population-state-only signals shows benefit in this regime
(Appendix~\ref{app:targeting-diagnostics}). Stratifying the 322 events
by problem difficulty separates two accounts of the low well-targeted
rate: on the hardest tercile, where the base-rate explanation least
applies, the well-targeted share rises only to $15.6\%$ (from $14.0\%$
overall) while high-flat-PRM stays dominant at $62.3\%$, so PRM flatness
rather than base-rate success is the binding constraint
(Appendix~\ref{app:difficulty-strat}).

\paragraph{Synthesis.} Every targeting threshold we tested either
admits cosmetic-plateau false positives
(Appendix~\ref{app:targeting-diagnostics}) or empties the injection
budget below the level at which any aggregate metric would move; the
three-axis novelty (PRM-pruned trigger, in-place sibling injection,
verbatim fragment) is operationally realized but produces no detectable
benefit. The random-rule control's matching parity at $2.4\times$ the
stagnation rate (\S\ref{sec:mechanism}, Table~\ref{tab:table1}) places
the inertness on the mechanism at this corner, not the particular
heuristic. \S\ref{sec:discussion} separates this from claims about the
broader fragment-grafting design space.

\paragraph{$\xi_{\min}$ sensitivity.} Sweeping $\xi_{\min}\in\{0.10,
0.30,0.50,0.70\}$ (Qwen/Math-Shepherd/MATH500), injection rate falls
monotonically from $16.0\%$ to $2.4\%$ (gate functional) yet no setting
moves Pass@$k$ or \modeRate outside the $n{=}3$ MDE band: the gate
rejects injections but is not the binding constraint
(Appendix~\ref{app:xi-sweep}).

\paragraph{Population size.} Because grafting is a population-level
mechanism, we sweep $N$ over a $32\times$ range ($N{=}8$ to $N{=}256$,
$n{=}100$ slice, 3 seeds/cell, matched populations). Parity holds at
every $N$ and never turns positive; the binding cell is $N{=}16$ Pass@8
at $-0.27$~pp, and at $N{=}256$ parity is essentially exact (indep
Pass@1 $0.7073$ vs.\ \ppfg-stag $0.7062$, seed-$\sigma \le 0.0014$).
Larger populations do not make the transfer useful
(Appendix~\ref{app:n-sweep}). This $n{=}100$ sweep is a distinct data
slice from Table~\ref{tab:table1} and has not yet been re-derived
under the \S\ref{sec:comparator-note} canonical comparator; we expect
the parity conclusion (a relative, within-sweep comparison) to be
robust to the correction, as it was for the $n{=}500$ cells, but have
not verified this for this slice specifically.

\paragraph{Compute matching.} At matched tokens-per-problem ($\sim4000$
for $N{=}8$ methods), BoN-$N{=}8$ and indep lead Pass@4 ($0.809$), SMC
$0.789$, \ppfg-stag $0.807$; at $2\times$ compute BoN-$N{=}16$ reaches
Pass@16$=0.891$ (pre-\S\ref{sec:comparator-note}-correction values; cf.\
Table~\ref{tab:table1} for the canonical Pass@4/Pass@16 figures on this
same $n{=}500$ population). \ppfg-stag never strictly dominates indep,
SMC, or BoN at any Pass@$k$ under either comparator
(Appendix~\ref{app:compute-match}, Figure~\ref{fig:compute-match}).

\paragraph{Ceiling-regime replication (GSM8K).} On GSM8K
\citep{gsm8k} at Pass@1 $\approx 0.87$ (Qwen2.5-7B-Instruct,
$n{=}500{\times}3$ seeds), indep, \ppfg-stag, and \ppfg-random produce
Pass@1 identical to three decimals (max $|\Delta|{=}0.03$~pp, within the
5-seed MDE $\pm1.4$~pp), Pass@8 within $0.0$~pp, \modeRate within
$0.27$~pp. Injection rates fall to $2.8\%$/$9.5\%$ (vs $16.0\%$/$28.8\%$
on MATH500) since chains rarely dip below $\tau_\mathrm{kill}{=}0.4$ on
arithmetic; the mechanism fires in proportion to task difficulty
(Appendix~\ref{app:gsm8k}).

\subsection{Mechanistic spot-check: post-injection PRM delta}
\label{sec:prm-delta}

For every \ppfg-stag injection event across the three Qwen2.5-7B
$n{=}500$ seeds ($N{=}322$), we compute the change in the target chain's
process-reward score around the injection step:
$\Delta = \overline{\mathrm{PRM}}_{[k,\,k+3)} - \overline{\mathrm{PRM}}_{[k-3,\,k)}$,
$k$ the grafting step index (windows bounded by chain end; no event skipped).

The mean $\Delta$ is $-0.134$ (median $-0.093$, IQR $[-0.215,\,-0.031]$);
272/322 events (84.5\%) have $\Delta < 0$. A Wilcoxon signed-rank test
against $H_0\!:\,\mathrm{median}(\Delta){=}0$ rejects decisively
($W{=}3497.5$, $p \approx 1\!\times\!10^{-40}$); Cliff's $\delta = -0.699$
(large effect). This sharpens rather than contradicts the inertness
verdict: the per-event PRM signal is not noisy around zero but reliably
depressed ($\sim$0.09--0.13 lower three steps after injection, despite
mean fragment quality $0.78\pm0.08$), consistent with the
regime-asymmetry account (\S\ref{sec:regime-asymmetry})---a donor
fragment that scored highly natively lands as a foreign-context stylistic
discontinuity the same PRM penalizes. This is inconsistent with the
charitable ``helps locally but averages out'' reading: grafts do not
appear to help locally either.

\paragraph{Chain dynamics after injection.} Bucketing the 322 events by
the injected chain's status three steps later, $25.8\%$ reach a pruned
terminal status vs.\ a $9.4\%$ matched-step baseline on non-injected
siblings (CI $[9.0,9.8]$), a $2.75\times$ rate ($2.59\times$ on the
289-event PRM-trajectory-matched subset, so the hazard is not pure
selection; Appendix~\ref{app:counterfactual}). A surviving-sibling
counterfactual on $65$ matched problem-seed pairs finds no detectable
compensation (no promotion lift $\Delta{=}{-}0.011$, pop-correctness
$\Delta{=}{+}0.015$, per-chain tokens slightly \emph{lower}
$\Delta{=}{-}20.2$, opposite a freed-budget account). The evidence is
consistent with a selection-dominated account: stagnation chains were
pre-selected for low/flat PRM and would have been pruned anyway;
\ppfg{} pulls pruning forward by $\approx 3$ steps without
redistributing completion.

\paragraph{Empty-bracket confound control.} To test whether the
$\Delta<0$ effect is the out-of-distribution bracket wrapping rather
than the foreign-style step content, we ran an empty-bracket \ppfg-stag
variant (identical targeting/extraction/gate; injected text replaced by
a content-free wrapper) at full MATH500 ($n{=}500$, 3 seeds). Over 241
events the mean $\Delta$ is $-0.0736$ ($W{=}4594$, $p\approx
5{\times}10^{-20}$; Cliff's $\delta=-0.515$), so the depression has both
a format and a content component; under (untested) additivity the
content-only term is $\Delta_{\mathrm{content}}=-0.060$, a $\approx
55\%/45\%$ split we read as bounded between format- and content-dominant
rather than a point estimate (\S\ref{sec:limitations} item~10). A
seamless variant (no wrapper) and a re-attention variant (post-graft
self-check) both leave Pass@$k$ within seed-$\sigma$ of independent, so
neither remedy alone unlocks a benefit here (full numbers:
Appendices~\ref{app:empty-bracket}, \ref{app:format-ablation}). The
residual content channel---the PRM mis-pruning grafted chains that would
otherwise have answered correctly, which would let a real benefit hide
behind flat Pass@$k$---is closed separately: granting injected chains a
3-step exemption from threshold pruning measurably reduces pruning, yet
graft content adds $-0.04$~pp at Pass@1 over a matched exemption-only
control with no spliced content (Appendix~\ref{app:prune-immunity}).
The elevated-hazard pattern itself replicates descriptively on both
further architectures ($2.18\times$, $1.83\times$;
Appendix~\ref{app:hazard-crossarch}).

% =========================================================================
\subsection{Replication across benchmarks (Qwen)}
\label{sec:qwen-benchmarks}

To establish inertness is not a MATH500 artifact, we replicate the
three-method sweep (Qwen2.5-7B-Instruct) on four further benchmarks:
AIME 2024+2025 ($n{=}120$, near-floor), NuminaMath-CoT
\citep{li2024numinamath} ($n{=}99$), GPQA-Diamond \citep{rein2023gpqa}
($n{=}198$, both Math-Shepherd and VersaPRM \citep{versaprm2025}), and
MMLU-Pro Law/Biology/Philosophy \citep{wang2024mmlupro} ($n{=}500$, both
PRMs). VersaPRM is a multi-domain LoRA adapter on Llama-PRM800K; a
held-out smoke check confirmed calibration on math ($+0.076$ gap) and
law ($+0.423$) before any end-to-end run.

The headline metrics across all four benchmarks are summarized in
Appendix~\ref{app:qwen-bench} (Table~\ref{tab:qwen-bench}), with full
9-cell per-benchmark tables. The
absolute capability range spans $0.0417$ (AIME) to $0.4116$ (Numina);
under VersaPRM, GPQA drops to $0.1109$ as the multi-domain PRM is a
stricter scorer at this depth. Across all four benchmarks $\times$ both
PRMs, every \ppfg-vs-independent contrast on \passat{1} and \modeRate
falls inside the per-benchmark MDE. The AIME ($0.0417$) and
VersaPRM-GPQA ($0.1109$) cells sit near the capability floor, so their
parity is \emph{corroborating} (\ppfg{} does not rescue a near-floor
regime) rather than \emph{discriminating}; the discriminating cells have
non-floor base rates (NuminaMath, Math-Shepherd-GPQA, MMLU-Pro, and the
MATH500 headline).

\paragraph{Cross-architecture replication on the extended benchmarks.}
LLaMA-3.1-8B-Instruct and DeepSeek-R1-Distill-Qwen-7B extend the
matrix on AIME, NuminaMath-CoT, and GPQA-Diamond (Math-Shepherd PRM);
cross-architecture replication on GSM8K and MMLU-Pro
Law/Biology/Philosophy is deferred (\S\ref{sec:limitations} item~3).
Every \ppfg-vs-indep contrast is inside the
per-benchmark MDE
(Appendix~\ref{app:cross-arch-newbench}).

% =========================================================================
\section{Discussion}
\label{sec:discussion}

\paragraph{Practical implications.}\label{sec:practical} For
parallel-CoT inference with a math-tuned PRM and a strong base LM,
Best-of-$N$ dominates Pass@$k$ at matched compute and \ppfg{} variants
are indistinguishable from independent on every cell tested.
\emph{Inert} here is population-level (no Pass@$k$ or \modeRate effect);
per-event the mechanism is mildly \emph{adverse}
(\S\ref{sec:prm-delta}: $-0.134$ PRM, $2.75\times$ hazard), a footprint
the selection account explains and the population absorbs.

\paragraph{Target for future falsification.}\label{sec:regime-asymmetry} The
configuration we characterize is the corner where none of the five
compensating ingredients prior work relies on (separate-solver-pass
re-attention, learned distillation, training-time gradient, verbal
summarization/style adaptation, cross-problem abstraction;
\S\ref{sec:related}) is present. \ppfg{}'s corner (PRM-pruned trigger
$\times$ in-place sibling $\times$ verbatim fragment $\times$ heuristic
targeting) pays for none; these ingredients were derived inductively
from those systems and are not exhaustive. Two added in isolation (the
seamless and re-attention variants, Appendix~\ref{app:format-ablation})
also stay within seed-$\sigma$ of independent, consistent with the
gains requiring more than a single ingredient.
Future work here should pre-register which ingredient it supplies and
test the prediction that a separate-solver-pass variant moves Pass@1.

\paragraph{A null benchmark future work can pre-register
against.}\label{sec:null-benchmark} This paper is diagnostic: it
characterizes a mechanism null and does not propose a positive method,
and we decline to manufacture one we do not have. What a diagnosis of
this kind can supply instead is a number to beat, and the oracle
analysis (Appendix~\ref{app:oracle-targeting}) is that number. A
hindsight oracle that gets to choose, per problem, between running
\ppfg-stag and running independent---an upper bound on any possible
per-problem deployment gate over this mechanism---exceeds the
independent baseline by at most $+0.13$~pp across
$k\in\{1,2,4,8\}$, inside seed-$\sigma$. Optimal deployment of bare
verbatim grafting buys essentially nothing, so whatever leverage exists
in this design space is not in deciding \emph{when} to graft.

Concretely, we propose two things of future inference-time
fragment-grafting work. (a)~Name explicitly which compensating
ingredient the design supplies, from the taxonomy above or beyond it.
(b)~Report against the oracle ceiling: a Pass@$k$ gain within
$+0.13$~pp of independent has not beaten optimal per-problem deployment
of this graft-content class, and a gain above it should be credited to
the ingredient named in (a) rather than to the transfer itself. The
bound is scoped---it covers per-problem deployment, not event-level
targeting (injecting only a hand-picked subset of events) and not
ingredient-side modifications---but within that scope it converts a
negative result into a pre-registrable comparison. Of the ingredients we
were able to test cheaply, the prompt-level ones do nothing: both
seamless rendering and a post-graft self-check directive stay within
seed-$\sigma$ (Appendix~\ref{app:format-ablation}), which points at the
heavier ingredients---training-time integration, cross-problem
abstraction, or a genuine separate solver pass---rather than at a
prompt-level fix. A separate-solver-pass re-attention variant in the
manner of \citet{sndt2026} is not ruled out by anything we report here
and is, on our reading, the most informative next experiment.

% =========================================================================
\section{Limitations}
\label{sec:limitations}

\begin{enumerate}\setlength{\itemsep}{0pt}\setlength{\parsep}{0pt}\setlength{\topsep}{2pt}
\item \textbf{Statistical power.} With three seeds per cell at $n=500$,
the minimum detectable effect at 80\% power (two-sided, $\alpha=0.05$)
is approximately 2.2 absolute points for \modeRate, 3.1 points for
Pass@8, and 1.8 points for Pass@1. The \S\ref{sec:results-main} and
\S\ref{sec:mechanism} ``indistinguishable from independent'' claims
should be read as ``no improvement larger than 2.2 points on \modeRate,
3.1 points on Pass@8, or 1.8 points on Pass@1 detectable.''
The four-bucket targeting analysis in
\S\ref{app:targeting-diagnostics} was originally hand-classified on the
first-100 slice (55 events); we have since reproduced it with the
identical deterministic classifier on the full $n=500$ corpus (322
events), with a consistent $14\%$ well-targeted rate
(\S\ref{app:targeting-diagnostics}), so the diagnostic is no longer
slice-bound. A higher-seed re-run would further tighten the aggregate
MDE bounds but is out of scope.

\item \textbf{PRM domain transfer.} Math-Shepherd is math-tuned with
Mistral-family training data; the \S\ref{sec:cross-arch} cells apply
it unchanged to LLaMA-3.1-8B-Instruct and DeepSeek-R1-Distill-Qwen-7B
outputs. The Table~\ref{tab:prm-dist} audit shows that absolute score
levels differ across architectures, but the prune-trigger rate at
$\tkill=0.4$ varies modestly (2.4\% / 4.3\% / 6.4\%), and the LLaMA
and DeepSeek cells actually fire \ppfg \emph{more} often than Qwen.
The cross-architecture parity is therefore not an artifact of the PRM
failing to trigger. \S\ref{sec:qwen-benchmarks} additionally
substitutes VersaPRM (\citealp{versaprm2025}; a multi-domain PEFT-LoRA
adapter on Llama-PRM800K) for Math-Shepherd on GPQA-Diamond and
MMLU-Pro Law/Biology/Philosophy and observes the same inertness pattern,
empirically ruling out ``the math-tuned PRM is the wrong scorer''
as an explanation of the negative finding at this regime. Absolute
accuracy on LLaMA-3.1-8B-Instruct and DeepSeek-R1-Distill-Qwen-7B in
Table~\ref{tab:cross-arch} is substantially depressed by scoring
censorship (roughly half of each population produces no extractable
answer and is scored incorrect by construction); see
\S\ref{sec:comparator-note} for the conditional-accuracy figures and
the answer-comparator correction underlying all numbers in this
revision.

\item \textbf{Base model coverage.} Experiments span Qwen2.5-7B-
Instruct, LLaMA-3.1-8B-Instruct, and DeepSeek-R1-Distill-Qwen-7B (the
latter is Qwen-family but distilled from a reasoning model). We note
that DeepSeek-R1-Distill-Qwen-7B shares its base LM lineage with
Qwen2.5-7B-Instruct; LLaMA-3.1-8B-Instruct is therefore the genuinely
cross-family replication, while DeepSeek-R1-Distill-Qwen-7B provides
a same-family reasoning-distilled replication at substantially
different absolute Pass@$k$. Mistral-7B-Instruct-v0.3 was attempted
but excluded. Generalization to non-Qwen, non-LLaMA architecture
families (Gemma, OLMo, Mixtral-of-Experts variants) untested.
Cross-architecture replication on GSM8K and MMLU-Pro
Law/Biology/Philosophy is untested in this submission; these two
benchmarks are reported on Qwen2.5-7B-Instruct only, so the
off-math-axis claim rests on Qwen alone for the two largest
non-math cells.

\item \textbf{Compute scale and injection density.} The headline
cells run \ppfg{} at $N=8$ ($N=16$ appears only for the Best-of-$N$
baseline). Behavior at larger populations is no longer untested: a
sweep to $N=256$ on the $n{=}100$ slice
(Appendix~\ref{app:n-sweep}) finds parity at every population size and
no delta exceeding $0.27$~pp in magnitude, with the effect never turning
positive, so the ``$N{=}8$ is too small for a population mechanism''
hypothesis is not supported across a $32\times$ range. That sweep runs
at $n{=}100$ rather than $n{=}500$, so it is a sensitivity analysis
rather than a second headline result, and populations beyond $N{=}256$
remain untested. The per-chain
injection cap $M=2$ was fixed at small-scale smoke testing and not
re-swept, so per-chain density remains a candidate binding constraint
the random-rule control (which raises per-\emph{population} density
$2.4\times$) does not fully close. We note, however, that raising
density is an implausible route to benefit given the per-event
footprint: each injection locally depresses PRM by $-0.134$ and
elevates the pruning hazard $2.75\times$ (\S\ref{sec:prm-delta}), so
applying \emph{more} of a locally-adverse graft is not expected to
help; an $M$ sweep would test this directly.

\item \textbf{English math, science, and graduate-MCQ only.} Benchmarks
span MATH500, AIME 2024+2025, NuminaMath-CoT, GSM8K (math),
GPQA-Diamond (graduate science), and MMLU-Pro Law/Biology/Philosophy
(mixed-domain MCQ). All are English-language with single-answer
ground-truth (numeric or letter). Open-ended generation, multilingual
reasoning, and tasks without a checkable gold answer are untested.

\item \textbf{Heuristic policy.} A learned target-selection policy is
future work.

\item \textbf{No human evaluation of fragment quality or graft
naturalness.} The \S\ref{sec:prm-delta} empty-bracket control and
the post-injection PRM-delta provide a verifier-side proxy for
graft naturalness, but human ratings of whether grafted text
reads as natural continuation versus stylistic break would test
the ``stylistic discontinuity'' interpretation of
\S\ref{sec:prm-delta} more directly; the present paper relies
on the PRM-side proxy and does not commit to human evaluation.

\item \textbf{Per-injection credit-assignment analysis} (PopMC-GRPO or
comparable MC value estimation) is not included in this submission;
the \S\ref{sec:mechanism} characterization rests on the three
lines of evidence enumerated in \S\ref{sec:intro} (parity, broadly
replicated; targeting-heuristic limitation, of which the false-positive
rate and compound-gate infeasibility are two views; and suggestive
mechanism-level evidence from the PRM-delta decomposition and the
surviving-sibling counterfactual), none of which substitutes for the
per-injection MC value estimate this item defers.

\item \textbf{Compatibility-gate $\xi_{\min}$} was set to $0.0$ in
\S\ref{sec:results-main}--\S\ref{sec:mechanism} results. The
four-threshold sweep in Appendix~\ref{app:xi-sweep}
(Table~\ref{tab:xi-sweep}) confirms inertness is preserved across
non-trivial gate settings (injection rate falls monotonically from
16.0\% at $\xi_{\min}=0$ to 2.4\% at $\xi_{\min}=0.70$ with no
Pass@$k$ or \modeRate delta outside the $n=3$ MDE band), but the
headline tables in \S\ref{sec:results-main} use the $\xi_{\min}=0.0$
configuration. The interpretation is that the gate's role in this
regime is to filter pathological inputs (zero-norm embeddings,
fragment-equals-target degenerate cases), not to alter the
mechanism's aggregate behavior.

\item \textbf{Mechanistic analysis} is limited. \S\ref{sec:prm-delta}
reports the post-vs-pre injection PRM delta distribution and the
empty-bracket format/content decomposition. The \emph{causal}
sub-claim---that injection locally accelerates pruning beyond selection,
resting on the $2.59\times$ matched-PRM-trajectory ratio
(Appendix~\ref{app:counterfactual})---is established on Qwen only. The
\emph{descriptive} elevated-hazard pattern does replicate on both
further architectures ($2.18\times$ on DeepSeek-R1-Distill-Qwen-7B,
$1.83\times$ on LLaMA-3.1-8B-Instruct, CIs non-overlapping from
baseline in both; Appendix~\ref{app:hazard-crossarch}), but a
descriptive replication of the ratio is not a replication of the
matched-PRM-trajectory control, and we do not treat it as one. Deeper mechanistic probes, attention-pattern analysis
around the spliced span, token-level logit comparison between
counterfactual injection and no-injection generations, are out of
scope for this submission.
Finer-grained ablations of the bracket markup---seamless (no markup)
and re-attention (self-check directive appended)---were run as
robustness checks; see Appendix~\ref{app:format-ablation}.

\item \textbf{Both PRMs may share a formatting sensitivity.} Two
independent controls bound the confound that the PRM penalizes spliced
\emph{formatting} rather than the transfer being inert: seamless
rendering removes all markup and leaves Pass@$k$ at parity
(Appendix~\ref{app:format-ablation}), and a prune-immunity arm with a
matched null-content control closes the content channel
(Appendix~\ref{app:prune-immunity}). Neither excludes the deeper
possibility that Math-Shepherd and VersaPRM \emph{both} react to
mid-chain in-context grafts as out-of-distribution text, in a way a
differently-trained scorer would not. Our PRMs are two of a small
family trained on broadly similar step-level supervision; a PRM-free
pruning criterion, or a scorer trained with spliced-context
augmentation, would be needed to exclude this, and we did not run one.

\item \textbf{Equivalence testing} (Appendix~\ref{app:tost-detail}) is
conducted at $n{=}5$ seeds per cell across all three architectures
(Qwen2.5-7B, LLaMA-3.1-8B, DeepSeek-R1-Distill-Qwen-7B), against bounds
pre-registered from the 3-seed MDE. Both one-sided nulls reject at
$\alpha=0.05$ on every Qwen and every LLaMA cell; on DSR1 only stag.\
Pass@1 reaches positive equivalence (random Pass@1 rejects only the
upper-side null and the four DSR1 Pass@8 / \modeRate cells remain
inconclusive because seed-level $\sigma$ on this architecture is
$2$--$3\times$ that of the other two). The positive-equivalence claim
therefore covers $13/18$ cross-arch cells outright; the four
inconclusive DSR1 cells are within-$\sigma$ but not formally rejected,
and we report them as such rather than widen the bound.
\end{enumerate}

% =========================================================================
\section{Ethics, broader impact}
\label{sec:ethics}

\ppfg is an inference-time mechanism evaluated on a fixed math-
reasoning benchmark; it introduces no new training data, no new model
weights, and no new annotated corpus. Its operational cost (a
sentence-embedding compatibility score plus an in-context
demonstration splice on a chain decoded by a model the system already
runs) is small relative to the parallel-CoT generation it conditions.
Because the mechanism characterization in \S\ref{sec:mechanism} finds
no detectable improvement at the configurations tested, there is no
behavioral capability newly introduced by the method that requires
safety review beyond what already attaches to the underlying base
LMs and to Math-Shepherd as a process reward model. The paper's
primary broader-impact contribution is methodological: a regime
characterization of a frequently-proposed mechanism class
(cross-trajectory step-level grafting at inference time), which we
hope will reduce the prevalence of speculative claims about
inference-time mechanism efficacy. We release the full code, configs, analysis
scripts (targeting classifier, TOST, oracle, hazard), CPU smoke tests,
and saved trajectories at \url{https://github.com/Khawaja-Murad/ppfg-anon}.

% =========================================================================
% BIB-FIX-IN-PROGRESS: a separate bibliography-cleanup pass is
% handling references resolution; do not hand-edit bibitems here.
\footnotesize
\setlength{\bibsep}{0.0ex plus 0.1ex}

% =========================================================================
\appendix
\section{Per-benchmark replication details}
\label{app:qwen-bench}

This appendix contains the full per-benchmark tables and prose
summarized in \S\ref{sec:qwen-benchmarks}.

\begin{table}[!ht]
\centering
\caption{Replication summary across four further benchmarks
(Qwen2.5-7B-Instruct, 3 seeds, indep vs \ppfg-stag; full per-benchmark
tables below). All $\Delta$ values within MDE.}
\label{tab:qwen-bench}
\resizebox{\linewidth}{!}{%
\begin{tabular}{lcccc}
\toprule
Benchmark (PRM) & indep $P{@}1$ & stag $P{@}1$ & $\Delta P{@}1$ & $\Delta$\modeRate \\
\midrule
AIME 2024+2025 (MS)  & 0.0417 & 0.0417 & $+0.00$ & $+1.60$ \\
NuminaMath (MS)      & 0.4116 & 0.4116 & $+0.00$ & $+1.02$ \\
GPQA-Diamond (MS)    & 0.2363 & 0.2353 & $-0.11$ & $-0.01$ \\
GPQA-Diamond (VP)    & 0.1109 & 0.1115 & $+0.06$ & $-0.86$ \\
MMLU-Pro L/B/P (MS)  & 0.3940 & 0.3937 & $-0.03$ & $-0.27$ \\
MMLU-Pro L/B/P (VP)  & 0.3247 & 0.3242 & $-0.05$ & $-0.10$ \\
\bottomrule
\end{tabular}}
\end{table}

\subsection{AIME 2024+2025 (very-hard math, near-floor regime)}

The mechanism characterization rests on inertness holding across the
capability range. AIME 2024+2025 (120 problems combined) places the
base model near the difficulty floor: Qwen2.5-7B-Instruct's
\passat{1} on this set is $0.04$ ($N{=}120$, 3 seeds; $\sim$15 chains
of 360 ever produce the gold integer). At this regime the question
``could PPFG lift the floor?'' is sharpest. Table~\ref{tab:aime}
reports the 9-cell sweep at the same operating point used elsewhere.
\ppfg-stag matches independent on \passat{1} to within $0.00$~pp
(matched to four decimals: $0.0417$ vs.\ $0.0417$), with \modeRate
$+1.60$~pp; the random control sits on the same frontier. We note that
\passat{1} parity at this floor is weak evidence on its own: the
estimator is deterministic on perfectly-failed problems, so AIME
corroborates that \ppfg{} does not \emph{rescue} a near-floor regime
but does not by itself discriminate inertness from task
unsolvability. We treat it as corroborating rather than as one of the
discriminating cells.

\begin{table}[!ht]
\centering
\caption{AIME 2024+2025 ($n{=}120$, 3 seeds, Qwen2.5-7B-Instruct).
MDE $\pm 2.4$~pp (\passat{1}) / $\pm 4.7$~pp (\modeRate). The
identical $0.0417$ across methods reflects $5$ problems
$\times\,3$ seeds $=15$ problem-seed instances solved
consistently at this floor regime; the unbiased
Chen-et-al-2021 \passat{1} estimator is deterministic on
perfectly-solved or perfectly-failed problems, so differences
between methods can only register on partial-success problems,
of which there are essentially none at \passat{1}$\approx 0.04$.}
\label{tab:aime}
\resizebox{\linewidth}{!}{%
\begin{tabular}{lccc}
\toprule
Method & Pass@1 & Pass@8 & Mode rate \\
\midrule
independent      & 0.0417 & 0.1111 & 0.1880 \\
\ppfg-stag       & 0.0417 & 0.1111 & 0.2040 \\
\ppfg-random     & 0.0417 & 0.1056 & 0.2137 \\
$\Delta$(stag $-$ indep)   & $+0.00$ & $+0.00$ & $+1.60$ \\
$\Delta$(random $-$ indep) & $+0.00$ & $-0.55$ & $+2.57$ \\
\bottomrule
\end{tabular}}
\end{table}

\subsection{NuminaMath-CoT (olympiad-tier mid-difficulty math)}

NuminaMath-CoT \citep{li2024numinamath} fills the difficulty point
between MATH500 (Qwen \passat{1}~$=0.64$) and AIME (Qwen
\passat{1}~$=0.04$): mixed olympiad / AOPS / competition sources, 99
test problems with gold answers parseable from $\backslash$boxed in
the reference solutions. \ppfg-stag matches independent on \passat{1}
to four decimals ($0.4116$); \modeRate $\Delta = +1.02$~pp, within MDE
($\pm 5.2$~pp).

\begin{table}[!ht]
\centering
\caption{NuminaMath-CoT test ($n{=}99$, 3 seeds, Qwen2.5-7B-Instruct).
MDE $\pm 2.6$~pp / $\pm 5.2$~pp.}
\label{tab:numina}
\resizebox{\linewidth}{!}{%
\begin{tabular}{lccc}
\toprule
Method & Pass@1 & Pass@8 & Mode rate \\
\midrule
independent      & 0.4116 & 0.6162 & 0.6655 \\
\ppfg-stag       & 0.4116 & 0.6128 & 0.6757 \\
\ppfg-random     & 0.4091 & 0.6094 & 0.6689 \\
$\Delta$(stag $-$ indep)   & $+0.00$ & $-0.34$ & $+1.02$ \\
$\Delta$(random $-$ indep) & $-0.25$ & $-0.68$ & $+0.34$ \\
\bottomrule
\end{tabular}}
\end{table}

\subsection{GPQA-Diamond (cross-PRM, off the math axis)}

GPQA-Diamond \citep{rein2023gpqa} extends the characterization off the
math axis: 198 expert-graduate-level MCQ across physics, biology, and
chemistry. We run the cross-PRM cell with both Math-Shepherd and
VersaPRM \citep{versaprm2025}. Under Math-Shepherd, Qwen2.5-7B's
\passat{1} is $0.2363$; under VersaPRM it drops to $0.1109$ (the
multi-domain PRM is the stricter scorer at this depth, but the
ordering of methods is preserved). All \ppfg{} contrasts within MDE
under both PRMs (Table~\ref{tab:gpqa-full}). The PRM-domain confound
is empirically ruled out at this regime.

\begin{table}[!ht]
\centering
\caption{GPQA-Diamond cross-PRM ($n{=}198$, 3 seeds,
Qwen2.5-7B-Instruct). MDE $\pm 1.9$~pp / $\pm 3.7$~pp.}
\label{tab:gpqa-full}
\resizebox{\linewidth}{!}{%
\begin{tabular}{lccc}
\toprule
Method & Pass@1 & Pass@8 & Mode rate \\
\midrule
\multicolumn{4}{l}{\emph{Math-Shepherd PRM}} \\
independent      & 0.2363 & 0.6044 & 0.6530 \\
\ppfg-stag       & 0.2353 & 0.5892 & 0.6529 \\
\ppfg-random     & 0.2348 & 0.5976 & 0.6554 \\
\midrule
\multicolumn{4}{l}{\emph{VersaPRM}} \\
independent      & 0.1109 & 0.3872 & 0.7650 \\
\ppfg-stag       & 0.1115 & 0.3923 & 0.7564 \\
\ppfg-random     & 0.1094 & 0.3737 & 0.7608 \\
\bottomrule
\end{tabular}}
\end{table}

\subsection{MMLU-Pro Law / Biology / Philosophy (stratified $n{=}500$)}

MMLU-Pro \citep{wang2024mmlupro} carries the regime characterization
fully off the math axis at the $n{=}500$ scale: $\approx 167$
problems each from law, biology, and philosophy (seed $=42$), three-
method sweep under both PRMs. Under Math-Shepherd, Qwen2.5-7B
\passat{1}~$=0.394$; under VersaPRM, $0.325$. All \ppfg{} contrasts
within MDE on \passat{1} and \modeRate under both PRMs
(Table~\ref{tab:mmlupro}).
Per-subject (biology/law/philosophy) Pass@1 deltas under
Math-Shepherd are all within $\pm 0.20$~pp of zero; inertness holds
per-subject.

\begin{table}[!ht]
\centering
\caption{MMLU-Pro Law/Biology/Philosophy stratified subset ($n{=}500$,
3 seeds).}
\label{tab:mmlupro}
\resizebox{\linewidth}{!}{%
\begin{tabular}{lccc}
\toprule
Method & Pass@1 & Pass@8 & Mode rate \\
\midrule
\multicolumn{4}{l}{\emph{Math-Shepherd PRM}} \\
independent      & 0.3940 & 0.6840 & 0.7213 \\
\ppfg-stag       & 0.3937 & 0.6853 & 0.7186 \\
\ppfg-random     & 0.3933 & 0.6847 & 0.7181 \\
\midrule
\multicolumn{4}{l}{\emph{VersaPRM}} \\
independent      & 0.3247 & 0.6340 & 0.7735 \\
\ppfg-stag       & 0.3242 & 0.6367 & 0.7725 \\
\ppfg-random     & 0.3200 & 0.6313 & 0.7676 \\
\bottomrule
\end{tabular}}
\end{table}

% =========================================================================
\FloatBarrier
\section{Targeting-policy diagnostics in detail}
\label{app:targeting-diagnostics}

This appendix contains the full four-bucket classification and the
compound-gate dry-run + end-to-end measurement summarized in
\S\ref{sec:mechanism}.

\subsection{Four-bucket false-positive classification}

All 55 \ppfg-stag injection events from the first-100 slice of the
$n{=}500$ Qwen corpus were classified along four mutually-non-exclusive
buckets defined as deterministic functions of saved trajectory data:
\textbf{target-succeeded} (target chain was scored correct),
\textbf{high-flat-PRM} (target's three-step pre-injection PRM had
mean $\geq 0.6$ and range $<0.05$), \textbf{near-completion} (target
produced its answer within two steps after injection), and
\textbf{plausibly-helped} (the complement of the union of the other
three). The classifier is deterministic and the bucket definitions are
reproduced by an independent re-implementation; we do not claim
inter-annotator agreement.

\begin{figure}[!ht]
\centering
\includegraphics[width=\columnwidth]{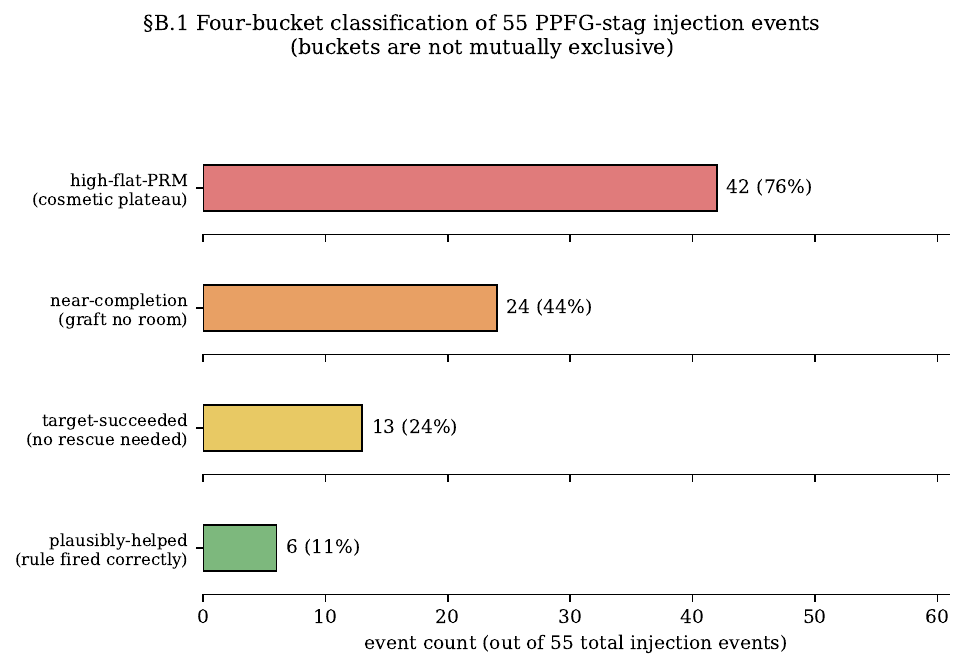}
\caption{Four-bucket classification of the 55 \ppfg-stag injection
events. Buckets are not mutually exclusive. Only $6/55$ ($11\%$) events
fall in \emph{plausibly-helped} (the complement of the union of the
other three); equivalently, $49/55$ ($89\%$) fall into at least one
state, \{high-flat-PRM, near-completion, target-succeeded\}, in which a
rescue graft cannot change the outcome. We report the $6/55$
well-targeted share as the primary quantity and the complement as the
share of wasted fires.}
\label{fig:buckets}
\end{figure}

\begin{table}[!ht]
\centering
\small
\resizebox{\columnwidth}{!}{%
\begin{tabular}{lrr}
\toprule
Bucket & Count & \% \\
\midrule
high-flat-PRM (cosmetic plateau)    & 42 & 76.4 \\
near-completion (graft no room)     & 24 & 43.6 \\
target-succeeded (no rescue needed) & 13 & 23.6 \\
\textbf{plausibly-helped (rule fired correctly)} & \textbf{6} & \textbf{10.9} \\
\bottomrule
\end{tabular}}
\caption{Four-bucket counts on the first-100 slice (55 events);
buckets are not mutually exclusive.}
\label{tab:buckets}
\end{table}

\paragraph{Full-corpus replication ($n{=}500$, 322 events).}
To verify the well-targeted rate is not an artifact of the
first-100 slice, we applied the identical deterministic classifier to
the full three-seed $n{=}500$ PPFG-stag corpus (322 injection events).
The well-targeted (plausibly-helped) share is $45/322 = 14.0\%$
(vs.\ $10.9\%$ on the slice); high-flat-PRM remains the dominant
disqualifier at $67.1\%$ (vs.\ $76.4\%$), with near-completion
$48.8\%$ and target-succeeded $18.9\%$. The picture is unchanged at
the larger scale: roughly seven in eight stagnation fires land where a
rescue graft cannot help, and the cosmetic-plateau bucket dominates.
The classifier is byte-identical to the slice analysis
(\texttt{scripts/reclassify\_buckets\_n500.py} reuses the
\texttt{inter\_annotator\_classify.py} event-gathering and
classification functions); output in
\texttt{results/bucket\_classification\_n500.json}.

\subsection{Difficulty-stratified targeting}
\label{app:difficulty-strat}

At an operating point where independent Pass@1 is $\approx 0.64$, the
``$86\%$ of fires do not need rescue'' statistic admits two readings that
the aggregate does not separate. Under a \emph{base-rate} account, the
target-succeeded and near-completion buckets are inflated mechanically
because the model already succeeds on most chains; under a
\emph{PRM-landscape} account, the binding constraint is that the PRM
score surface is flat and high even on chains that are struggling, so
population-state signals cannot identify a rescue target. The two make
opposite predictions on hard problems, where base-rate success is low:
under the base-rate account the well-targeted share should rise
substantially; under the PRM-landscape account it should not.

We stratify all 322 events by problem difficulty. Terciles are built from
pooled independent-baseline Pass@1 (24 chains per problem, seeds
42--44, fixed extractor), so the stratification is computed
independently of the \ppfg{} population being classified. Event
gathering and bucket classification are unmodified, so totals reconcile
against the full-corpus numbers above.

\begin{table}[!ht]
\centering \small
\caption{Targeting quality by problem-difficulty tercile (322
\ppfg-stag events, Qwen2.5-7B-Instruct + MATH500 $n{=}500$, 3 seeds).
Terciles are defined by pooled independent-baseline Pass@1. The
stagnation rule concentrates firing on hard problems, but the
well-targeted share barely rises there, while high-flat-PRM remains
dominant.}
\label{tab:difficulty-strat}
\resizebox{\linewidth}{!}{%
\begin{tabular}{lrrrr}
\toprule
Tercile & mean indep & $n$ events & well- & high-flat- \\
        & Pass@1     &            & targeted & PRM \\
\midrule
hard   & $0.108$ & $212$ ($66\%$) & $15.6\%$ & $62.3\%$ \\
medium & $0.721$ & $102$ ($32\%$) & $11.8\%$ & $74.5\%$ \\
easy   & $0.973$ & $8$ ($2\%$)    & $0.0\%$  & $100.0\%$ \\
\midrule
\emph{all} & --- & $322$ & $14.0\%$ & $67.1\%$ \\
\bottomrule
\end{tabular}}
\end{table}

\paragraph{Results.} Two things follow from
Table~\ref{tab:difficulty-strat}. First, the stagnation rule is not
wasting its budget on easy problems: $66\%$ of events land in the
hardest tercile, and only $2\%$ in the easiest. Second, and bearing
directly on the two accounts, restricting to the hardest tercile---where
base-rate inflation least applies---the well-targeted share rises only
to $15.6\%$, against $14.0\%$ over the full corpus. That is the
direction a base-rate account predicts, but far short of its magnitude:
if the null were primarily a base-rate artifact, the hard tercile should
show a materially higher well-targeted rate. Meanwhile high-flat-PRM
remains the dominant disqualifier at $62.3\%$ on the hardest problems,
outweighing well-targeting roughly $4{:}1$. This makes precise the
pattern already visible in the first-100-versus-full-500 split ($11\%$
vs.\ $14\%$): PRM flatness, not base-rate success, is the binding
constraint on targeting quality in this regime. Output in
\texttt{results/targeting\_by\_difficulty\_stratum.json}; code in
\texttt{scripts/targeting\_by\_difficulty\_stratum.py}.

\subsection{Compound-gate refinement (dry-run + end-to-end)}

The compound version additionally required (a) target mean recent PRM
below $\tau_\mathrm{low}$, (b) $\geq 4$ steps before $T_\mathrm{max}$,
(c) target's latest PRM below the median of active siblings, and (d)
$\geq 3$ active siblings. A dry-run on the 55 events at
$\tau_\mathrm{low}\in \{0.60, 0.70\}$ admits $9.1\%$ and $16.4\%$
respectively. The end-to-end measurement at $n{=}500{\times}3$ seeds
confirms the dry-run: both thresholds produce deltas within seed-$\sigma$
of independent (Table~\ref{tab:table3b}).

\begin{table*}[!ht]
\centering \small
\renewcommand{\arraystretch}{0.92}
\resizebox{\textwidth}{!}{%
\begin{tabular}{lrrrrrrr}
\toprule
Method & Pass@1 & Pass@4 & Pass@8 & \modeRate $\downarrow$ & \distinct $\uparrow$ & tpp & inj.\ rate \\
\midrule
independent ($N{=}8$)                              & $0.644 \pm .006$ & $0.809 \pm .009$ & $0.852 \pm .009$ & $0.789 \pm .006$ & $0.535 \pm .003$ & 4055 & --- \\
\ppfg-stag ($N{=}8$)                                & $0.643 \pm .006$ & $0.807 \pm .009$ & $0.849 \pm .011$ & $0.790 \pm .008$ & $0.535 \pm .003$ & 4048 & 0.215 (322/1500) \\
\textbf{\ppfg-stag-compound ($t{=}0.60$, $N{=}8$)} & $0.649 \pm .006$ & $0.810 \pm .008$ & $0.855 \pm .007$ & $0.790 \pm .006$ & $0.536 \pm .003$ & 3974 & \textbf{0.021 (32/1500)} \\
\textbf{\ppfg-stag-compound ($t{=}0.70$, $N{=}8$)} & $0.649 \pm .006$ & $0.810 \pm .008$ & $0.855 \pm .007$ & $0.791 \pm .005$ & $0.536 \pm .003$ & 3971 & \textbf{0.037 (56/1500)} \\
\bottomrule
\end{tabular}}
\caption{End-to-end measurement of the compound-gate variant at two
thresholds (3 seeds $\times$ 500 problems each). Pre-comparator-correction
values (\S\ref{sec:comparator-note}); the within-table parity finding
is not expected to change.}
\label{tab:table3b}
\end{table*}

% =========================================================================
% (No FloatBarrier here: Table 9 is the only spanning float pending, and
% Section C has none of its own -- letting the barrier land after Table 10
% instead lets C's heading/prose backfill the previous page instead of
% forcing an empty-column jump.)
\section{GSM8K replication (full table)}
\label{app:gsm8k}

To rule out a MATH500-specific artifact, we replicate at the same
operating point (Qwen2.5-7B-Instruct, $n{=}500$, 3 seeds, $N{=}8$) on
GSM8K \citep{gsm8k}, intentionally the easier task (Pass@1 $\approx
0.87$, near ceiling), where the strong-base-LM caveat
(\S\ref{sec:limitations} item~1) is most acute. indep, \ppfg-stag, and
\ppfg-random produce Pass@1 identical to three decimals (max
$|\Delta|{=}0.03$~pp, within the 5-seed MDE $\pm 1.4$~pp). The
proportional drop in injection rate ($2.8\%$ / $9.5\%$ vs $16.0\%$ /
$28.8\%$ on MATH500) is itself characterization: the mechanism fires
in proportion to how often the PRM identifies pruneable chains, which
is itself a function of task difficulty.

\begin{table}[!ht]
\centering \small
\begin{tabular}{lrrrr}
\toprule
 Method & inj.\ rate & P@1 & P@8 & \modeRate \\
\midrule
 indep        & ---               & $.871_{\pm.007}$ & $.958_{\pm.003}$ & $.959_{\pm.004}$ \\
 \ppfg-stag   & $2.8_{\pm 1.0}\%$ & $.871_{\pm.007}$ & $.958_{\pm.003}$ & $.958_{\pm.005}$ \\
 \ppfg-random & $9.5_{\pm 1.4}\%$ & $.871_{\pm.007}$ & $.958_{\pm.003}$ & $.961_{\pm.004}$ \\
\bottomrule
\end{tabular}
\caption{GSM8K replication ($n{=}500$, 3 seeds; tpp $\approx 1891$ for
all rows).}
\label{tab:gsm8k}
\end{table}

% =========================================================================
% Section order swapped (formatting pass): the extended cross-arch table
% opens directly on a spanning table* with only one short intro sentence,
% so it left the rest of the page blank. The empty-bracket section is
% pure prose (no floats) and backfills that page instead.
\FloatBarrier
\section{Empty-bracket confound control (full per-event detail)}
\label{app:empty-bracket}

Over 241 empty-bracket injection events across three seeds, the
per-event $\Delta$ distribution has mean $-0.0736$ (median $-0.0434$,
IQR $[-0.133, -0.005]$, std $0.1125$); 182/241 events ($75.5\%$) have
$\Delta < 0$ and 58 ($24.1\%$) have $\Delta > 0$. Wilcoxon signed-rank
against $H_0\!:\,\mathrm{median}(\Delta){=}0$ rejects ($W{=}4594$,
$p\approx 5{\times}10^{-20}$); Cliff's $\delta = -0.515$ (large
effect). $\Delta_{\mathrm{stag}}$ and $\Delta_{\mathrm{empty}}$ are
each individually significant ($p < 10^{-19}$). The derived
$\Delta_{\mathrm{content}} = \Delta_{\mathrm{stag}} -
\Delta_{\mathrm{empty}} = -0.060$ is presented under the additivity
assumption stated in \S\ref{sec:prm-delta} and bounded between
format-dominant and content-dominant readings as discussed there;
we do not assign $\Delta_{\mathrm{content}}$ its own significance
test in this submission. Under that assumption a portion of the
post-injection PRM depression is attributable to the
bracketed-demonstration wrapping that any in-context graft would
carry, and the rest to spliced step content scored as stylistically
discontinuous by the PRM; consistent with the main-text reading we
treat this only as bounded between format-dominant and
content-dominant, not as a $\approx$50/50 point estimate.

The empty-bracket event count (241) is lower than \ppfg-stag (322)
at matched $n$ and rule because population state diverges after the
first injection, so subsequent targeting decisions differ; the rule
definition is identical but the population-state context at each step
is not.

% =========================================================================
\FloatBarrier
\section{Cross-architecture replication on the extended benchmarks (full table)}
\label{app:cross-arch-newbench}

Full $3\times$-benchmark $\times$ 2-architecture sweep summarized in
\S\ref{sec:qwen-benchmarks}; Math-Shepherd PRM throughout, 3 seeds
per cell.

\begin{table*}[!ht]
\centering
\renewcommand{\arraystretch}{0.9}
\resizebox{\linewidth}{!}{%
\begin{tabular}{llcccccc}
\toprule
Arch & Method & Pass@1 & Pass@2 & Pass@4 & Pass@8 & Mode rate & $\Delta$P@1 vs indep \\
\midrule
\multicolumn{8}{l}{\emph{AIME 2024+2025 ($n{=}120$)} \quad MDE $\pm 2.4 / 4.7$~pp} \\
LLaMA-3.1-8B-Instruct & independent  & 0.0108 & 0.0186 & 0.0297 & 0.0444 & 0.6020 & --- \\
                      & \ppfg-stag   & 0.0101 & 0.0175 & 0.0282 & 0.0417 & 0.6219 & $-0.07$ \\
                      & \ppfg-random & 0.0094 & 0.0161 & 0.0256 & 0.0389 & 0.6534 & $-0.14$ \\
DeepSeek-R1-Distill-Qwen-7B & independent  & 0.0094 & 0.0173 & 0.0301 & 0.0500 & 0.6647 & --- \\
                            & \ppfg-stag   & 0.0090 & 0.0166 & 0.0287 & 0.0472 & 0.6764 & $-0.04$ \\
                            & \ppfg-random & 0.0097 & 0.0179 & 0.0309 & 0.0500 & 0.6521 & $+0.03$ \\
\midrule
\multicolumn{8}{l}{\emph{NuminaMath-CoT ($n{=}99$)} \quad MDE $\pm 2.6 / 5.2$~pp} \\
LLaMA-3.1-8B-Instruct & independent  & 0.2538 & 0.3203 & 0.3796 & 0.4343 & 0.6626 & --- \\
                      & \ppfg-stag   & 0.2525 & 0.3181 & 0.3765 & 0.4310 & 0.6676 & $-0.13$ \\
                      & \ppfg-random & 0.2504 & 0.3154 & 0.3717 & 0.4209 & 0.6500 & $-0.34$ \\
DeepSeek-R1-Distill-Qwen-7B & independent  & 0.2256 & 0.3048 & 0.3812 & 0.4613 & 0.6073 & --- \\
                            & \ppfg-stag   & 0.2226 & 0.2998 & 0.3746 & 0.4545 & 0.6050 & $-0.30$ \\
                            & \ppfg-random & 0.2260 & 0.3065 & 0.3855 & 0.4680 & 0.6257 & $+0.04$ \\
\midrule
\multicolumn{8}{l}{\emph{GPQA-Diamond ($n{=}198$)} \quad MDE $\pm 1.9 / 3.7$~pp} \\
LLaMA-3.1-8B-Instruct & independent  & 0.1023 & 0.1720 & 0.2687 & 0.3973 & 0.6001 & --- \\
                      & \ppfg-stag   & 0.0955 & 0.1599 & 0.2485 & 0.3670 & 0.6154 & $-0.68$ \\
                      & \ppfg-random & 0.0911 & 0.1516 & 0.2329 & 0.3367 & 0.6347 & $-1.12$ \\
DeepSeek-R1-Distill-Qwen-7B & independent  & 0.1370 & 0.2237 & 0.3348 & 0.4646 & 0.5903 & --- \\
                            & \ppfg-stag   & 0.1345 & 0.2216 & 0.3346 & 0.4663 & 0.5997 & $-0.25$ \\
                            & \ppfg-random & 0.1271 & 0.2081 & 0.3127 & 0.4377 & 0.5932 & $-0.99$ \\
\bottomrule
\end{tabular}}
\caption{Cross-architecture sweep on extended benchmarks. All
contrasts within per-benchmark MDE.}
\label{tab:cross-arch-newbench}
\end{table*}

% =========================================================================
% Section order swapped (formatting pass): the algorithm+diagram section
% opens with a full-width figure and has no leading prose of its own, so
% putting it right after Table 11 left an entire empty second column on
% that page. Putting the pure-prose empty-bracket section second instead
% lets it backfill whatever column is left after the algorithm/diagram,
% the way it previously backfilled the page before Table 11.
\FloatBarrier
\section{PPFG algorithm + architectural diagram}
\label{app:algorithm}

\begin{figure*}[!ht]
\centering
\includegraphics[width=0.95\textwidth]{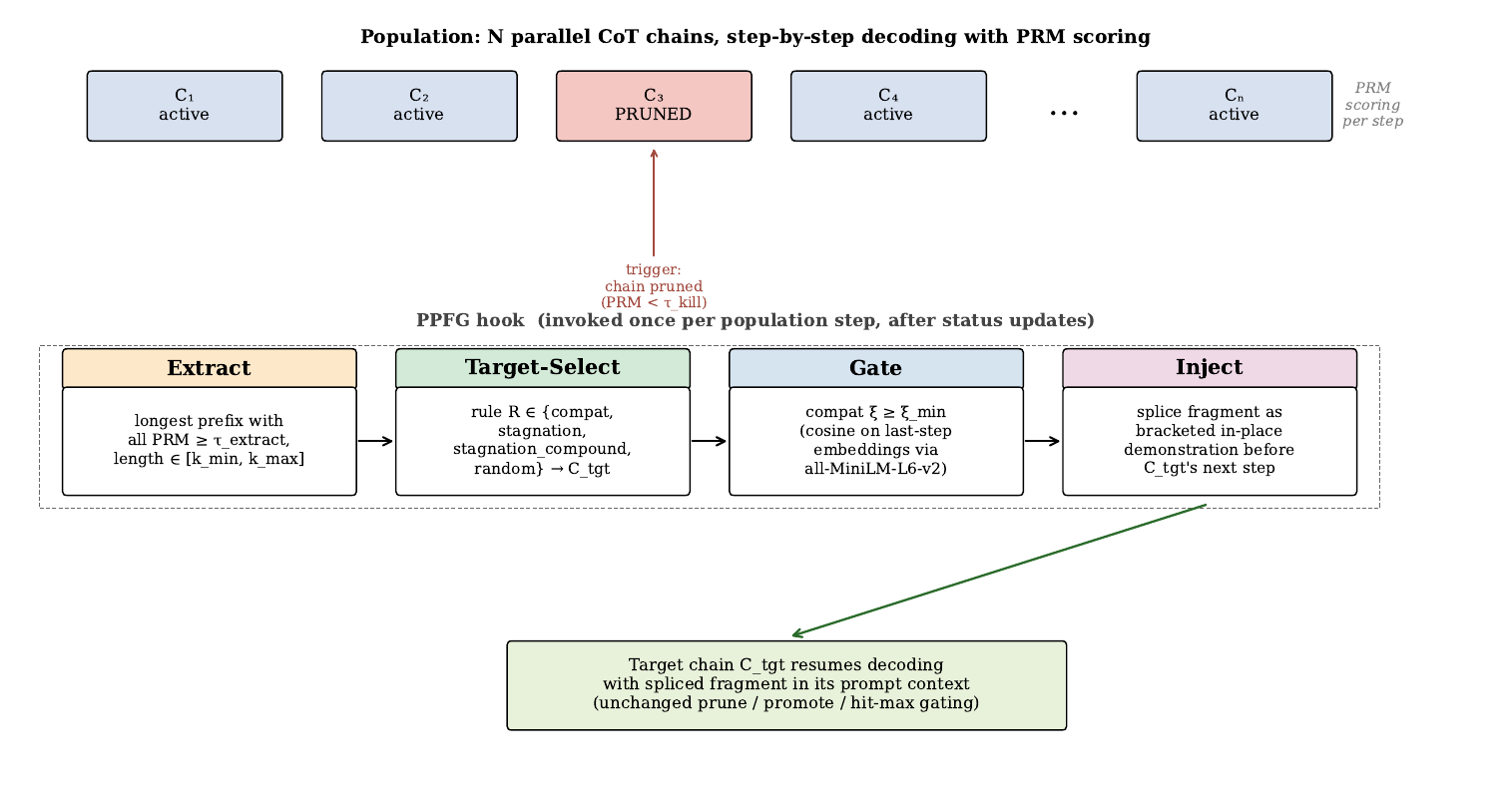}
\caption{The \ppfg mechanism. When a chain is pruned by the PRM (red
status box), its high-PRM prefix is extracted; the four \ppfg
sub-operations (extract / target-select / gate / inject) select a
still-decoding sibling chain and splice the fragment into the
sibling's prompt context as a bracketed in-place demonstration.}
\label{fig:arch}
\end{figure*}

\begin{algorithm}[H]
\caption{\ppfg in a parallel-CoT population.}
\label{alg:ppfg}
\small
\begin{algorithmic}[1]
\Require problem $q$, base model $\pi_\theta$, PRM $\rho$, sentence
embedder $\phi$, params $(N, \tkill, \textract, k_\mathrm{min},
k_\mathrm{max}, M, \xi_\mathrm{min}, R, T_\mathrm{max})$
\Ensure final answer (or $\bot$)
\State init $P_0 \gets N$ chains with prompt $q$, status $=$ ACTIVE
\For{$t = 1, 2, \dots, T_\mathrm{max}$}
  \State $A_t \gets \{C \in P_t : C.\mathit{status} = \mathrm{ACTIVE}\}$
  \If{$A_t = \emptyset$} \textbf{break} \EndIf
  \For{$C \in A_t$ in batch}
    \State $C.\mathit{append}(\mathrm{sample}\ s_t \sim \pi_\theta)$
    \State $C.\mathit{prm\_scores}.\mathit{append}(\rho(q, C.\mathit{steps}))$
  \EndFor
  \State $\mathit{newly\_pruned} \gets [\,]$
  \For{$C \in A_t$}
    \If{$\mathit{detect\_final\_answer}(C.\mathit{steps}[-1]) \neq \bot$}
      $C.\mathit{status} \gets \mathrm{PROMOTED}$
    \ElsIf{$C.\mathit{prm\_scores}[-1] < \tkill$}
      $C.\mathit{status} \gets \mathrm{PRUNED}$;
      $\mathit{newly\_pruned}.\mathit{add}(C)$
    \ElsIf{$|C.\mathit{steps}| \geq T_\mathrm{max}$}
      $C.\mathit{status} \gets \mathrm{HIT\_MAX}$
    \EndIf
  \EndFor
  \For{$C_\mathrm{src} \in \mathit{newly\_pruned}$} \Comment{PPFG hook}
    \State $F \gets$ longest prefix of $C_\mathrm{src}$ with all PRM
      $\geq \textract$, length $\in [k_\mathrm{min}, k_\mathrm{max}]$
    \If{$F = \bot$} \textbf{continue} \EndIf
    \State $C_\mathrm{tgt} \gets \mathit{select\_target}(F, \mathit{
      candidates}(A_t, M), R)$
    \If{$C_\mathrm{tgt} = \bot$} \textbf{continue} \EndIf
    \If{$\xi(F, C_\mathrm{tgt}) < \xi_\mathrm{min}$} \textbf{continue}
      \EndIf
    \State inject $F$ into $C_\mathrm{tgt}$'s prompt
  \EndFor
\EndFor
\State $\mathit{ans} \gets \{C.\mathit{final\_answer} : C.\mathit{status} = \mathrm{PROMOTED}\}$
\State \Return $\mathit{majority\_vote\_or\_highest\_mean\_prm}(\mathit{ans})$
\end{algorithmic}
\end{algorithm}

% =========================================================================
\section{Cross-architecture replication on MATH500 (full table + PRM-distribution audit)}
\label{app:cross-arch}

\begin{figure*}[!ht]
\centering
\includegraphics[width=0.95\textwidth]{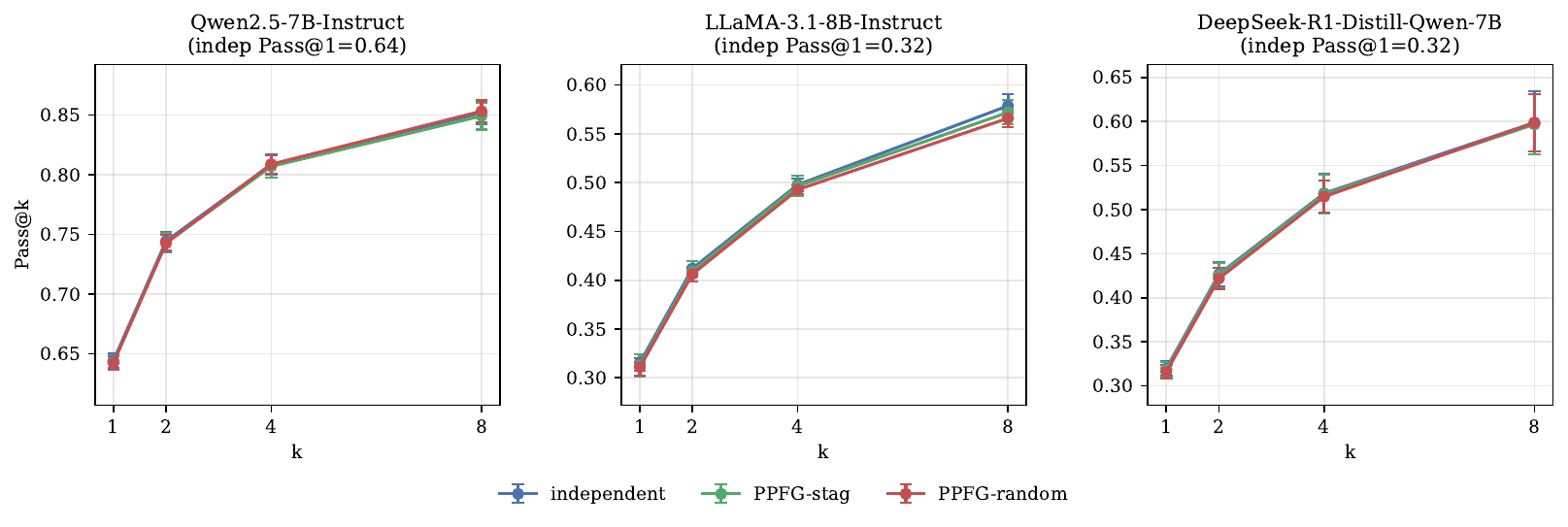}
\caption{Cross-architecture parity panel. Pass@$k$ vs $k$ on full
MATH500 ($n=500$, 3 seeds, mean $\pm$ seed-$\sigma$) for three base
architectures. \ppfg-stag and \ppfg-random sit within seed-$\sigma$ of
the independent baseline at every $k$ on every architecture.}
\label{fig:cross-arch}
\end{figure*}

\S\ref{sec:cross-arch} summarizes the \ppfg-stag-vs-independent Pass@1
deltas for LLaMA-3.1-8B-Instruct and DeepSeek-R1-Distill-Qwen-7B.
Table~\ref{tab:prm-dist} gives the PRM-score-distribution audit
underlying that section's ``PRM never prunes'' check;
Figure~\ref{fig:cross-arch} plots the parity panel; and
Table~\ref{tab:cross-arch} gives the full Pass@$k$/\modeRate/%
\distinct/tpp breakdown per architecture and method underlying both.

\begin{table}[!ht]
\centering \small
\begin{tabular}{lrrrrr}
\toprule
Arch. & $n$ scores & mean & q25 & q75 & $\tkill$ fire \\
\midrule
Qwen     & 89{,}091  & 0.768 & 0.624 & 0.917 & 2.4\% \\
LLaMA    & 112{,}022 & 0.723 & 0.593 & 0.881 & 4.3\% \\
DeepSeek &  90{,}323 & 0.715 & 0.562 & 0.893 & 6.4\% \\
\bottomrule
\end{tabular}
\caption{PRM-distribution audit: \ppfg{} fires \emph{more} often on
LLaMA and DeepSeek than on Qwen, ruling out a ``PRM never prunes''
artifact.}
\label{tab:prm-dist}
\end{table}

\FloatBarrier

\begin{table*}[!ht]
\centering \small
\renewcommand{\arraystretch}{0.9}
\resizebox{\textwidth}{!}{%
\begin{tabular}{llrrrrrrr}
\toprule
Architecture & Method & Pass@1 & Pass@2 & Pass@4 & Pass@8 & \modeRate $\downarrow$ & \distinct $\uparrow$ & tpp \\
\midrule
\multirow{3}{*}{Qwen2.5-7B-Instruct} & independent  & $0.592 \pm .007$ & $0.683 \pm .006$ & $0.742 \pm .005$ & $0.783 \pm .006$ & $0.788 \pm .009$ & $0.535 \pm .003$ & 4030 \\
                                      & \ppfg-stag   & $0.591 \pm .008$ & $0.682 \pm .007$ & $0.742 \pm .007$ & $0.783 \pm .011$ & $0.789 \pm .010$ & $0.535 \pm .003$ & 4025 \\
                                      & \ppfg-random & $0.591 \pm .007$ & $0.682 \pm .006$ & $0.743 \pm .005$ & $0.786 \pm .007$ & $0.788 \pm .010$ & $0.535 \pm .002$ & 4019 \\
\midrule
\multirow{3}{*}{LLaMA-3.1-8B-Instruct} & independent  & $0.316 \pm .007$ & $0.415 \pm .007$ & $0.504 \pm .007$ & $0.584 \pm .011$ & $0.606 \pm .012$ & $0.555 \pm .006$ & 3629 \\
                                       & \ppfg-stag   & $0.313 \pm .011$ & $0.409 \pm .011$ & $0.497 \pm .010$ & $0.572 \pm .013$ & $0.612 \pm .009$ & $0.553 \pm .005$ & 3595 \\
                                       & \ppfg-random & $0.311 \pm .009$ & $0.406 \pm .007$ & $0.493 \pm .005$ & $0.566 \pm .009$ & $0.596 \pm .007$ & $0.551 \pm .006$ & 3601 \\
\midrule
\multirow{3}{*}{DeepSeek-R1-Distill-Qwen-7B} & independent  & $0.314 \pm .013$ & $0.418 \pm .016$ & $0.508 \pm .022$ & $0.586 \pm .031$ & $0.664 \pm .013$ & $0.634 \pm .004$ & 2001 \\
                                              & \ppfg-stag   & $0.318 \pm .009$ & $0.425 \pm .015$ & $0.518 \pm .023$ & $0.597 \pm .034$ & $0.667 \pm .010$ & $0.633 \pm .004$ & 2045 \\
                                              & \ppfg-random & $0.316 \pm .008$ & $0.422 \pm .013$ & $0.515 \pm .021$ & $0.599 \pm .032$ & $0.651 \pm .010$ & $0.634 \pm .005$ & 2052 \\
\bottomrule
\end{tabular}}
\caption{Cross-architecture results on full MATH500 ($n=500$), mean
$\pm$ std, canonical scoring path (\S\ref{sec:comparator-note}).
\emph{independent} rows: 5 seeds ($\{42,\ldots,46\}$); LLaMA/DeepSeek
\ppfg\ rows: 3 seeds (a seed-45/46 relaunch ambiguity for those cells
is unresolved). LLaMA/DeepSeek's low absolute Pass@1 is substantially
a censoring effect, not weaker per-answer accuracy---see
\S\ref{sec:comparator-note}.}
\label{tab:cross-arch}
\end{table*}

% =========================================================================
\FloatBarrier
\section{Compute matching figure}
\label{app:compute-match}

\begin{figure}[!ht]
\centering
\includegraphics[width=\columnwidth]{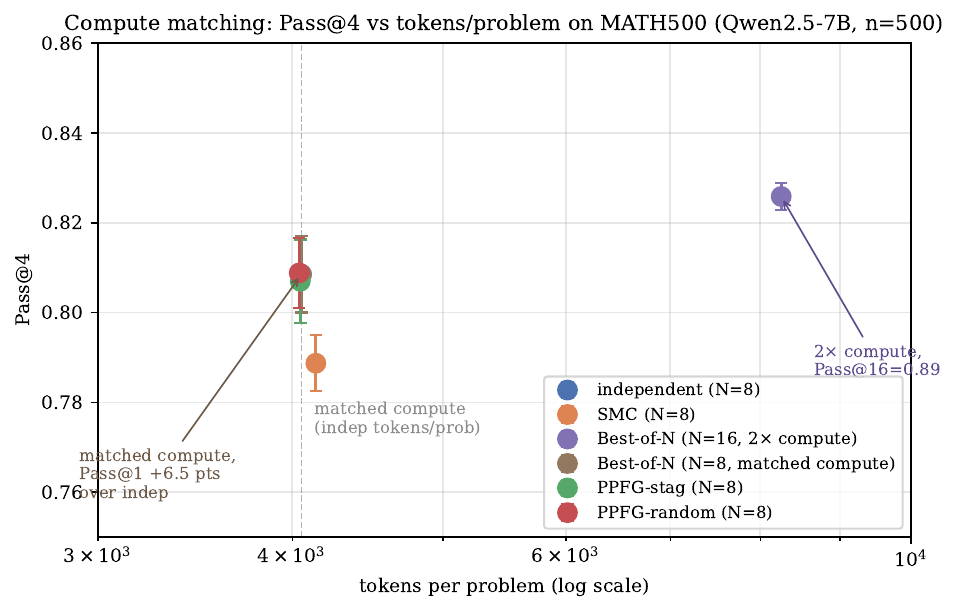}
\caption{Pass@4 vs tokens-per-problem on MATH500 (Qwen2.5-7B,
$n{=}500$, log-$x$). BoN at matched compute dominates the matched-tpp
column; BoN at $N{=}16$ reaches Pass@16$=0.89$ at $2\times$ compute.}
\label{fig:compute-match}
\end{figure}

% =========================================================================
\FloatBarrier
\section{$\xi_{\min}$ sensitivity sweep (full table)}
\label{app:xi-sweep}

\S\ref{sec:mechanism} summarizes this sweep: injection rate falls
monotonically as $\xi_{\min}$ rises, yet no setting moves Pass@$k$ or
\modeRate outside the MDE band. Table~\ref{tab:xi-sweep} below gives
the full per-threshold breakdown.

\begin{table*}[!ht]
\centering \small
\setlength{\tabcolsep}{6pt}
\renewcommand{\arraystretch}{0.9}
\begin{tabular}{lrrrrrrr}
\toprule
 $\xi_{\min}$ & inj\_rate & Pass@1 & Pass@4 & Pass@8 & \modeRate & \distinct & tpp \\
\midrule
 0.00 & 16.0$\pm$1.1 & 64.3$\pm$0.5 & 80.7$\pm$0.8 & 84.9$\pm$0.9 & 79.0$\pm$0.7 & 53.5$\pm$0.2 & 4048 \\
 0.10 & 12.5$\pm$0.7 & 64.7$\pm$0.5 & 80.9$\pm$0.7 & 85.3$\pm$0.7 & 78.7$\pm$0.6 & 53.6$\pm$0.2 & 3967 \\
 0.30 & 11.0$\pm$0.3 & 64.8$\pm$0.5 & 80.9$\pm$0.7 & 85.3$\pm$0.7 & 78.9$\pm$0.6 & 53.6$\pm$0.2 & 3970 \\
 0.50 &  7.0$\pm$0.3 & 64.8$\pm$0.5 & 80.9$\pm$0.7 & 85.3$\pm$0.7 & 78.7$\pm$0.5 & 53.6$\pm$0.2 & 3974 \\
 0.70 &  2.4$\pm$0.7 & 64.8$\pm$0.5 & 80.9$\pm$0.7 & 85.3$\pm$0.7 & 78.7$\pm$0.6 & 53.6$\pm$0.2 & 3973 \\
\bottomrule
\end{tabular}
\caption{$\xi_{\min}$ sensitivity sweep on Qwen2.5-7B + MATH500
($n=500$, 3 seeds per cell). Values in percentage points except tpp.
Pre-comparator-correction values (\S\ref{sec:comparator-note}); the
within-column null finding is not expected to change.}
\label{tab:xi-sweep}
\end{table*}

% =========================================================================
\FloatBarrier
\section{TOST equivalence: methodology and per-cell discussion}
\label{app:tost-detail}

The Table~\ref{tab:table1} \ppfg-vs-independent deltas all sit within
one seed standard error of zero, but a within-$1\sigma$ window is not
by itself an equivalence test: a reviewer is entitled to ask whether
the data also rule out a meaningful hidden effect just below the
variance floor. We address this with two one-sided Welch $t$-tests
(TOST) at $\alpha=0.05$ against \emph{pre-registered} bounds equal to
the 3-seed minimum detectable effect sizes from
\S\ref{sec:limitations} item 1, $\pm 0.018$ for Pass@1,
$\pm 0.031$ for Pass@8, and $\pm 0.022$ for the answer-mode rate.
These bounds were locked before any TOST $p$-value was computed.

On the Qwen rows the 5-seed procedure positively rejects \emph{both}
one-sided nulls at $\alpha=0.05$ on every metric for both injection
rules: at the pre-registered bounds, the data are statistically
inconsistent with \ppfg{} hiding a positive or negative effect of more
than $\pm 1.8$ points on Pass@1, $\pm 3.1$ on Pass@8, or $\pm 2.2$ on
\modeRate. With the cross-architecture cells also extended to 5 seeds
($\{42,\ldots,46\}$), all six LLaMA-3.1-8B cells (stag.\ and random,
both Pass@$k$ and \modeRate) now likewise reject \emph{both} one-sided
nulls and return ``equivalent'' verdicts. The DSR1-Qwen-7B cells remain
the noisiest: stag.\ Pass@1 reaches equivalence at 5 seeds and random
Pass@1 rejects only the upper-side null (mean $\Delta=-0.36$~pp), but
the four Pass@8 and \modeRate cells stay inconclusive because the
seed-to-seed standard deviations on DSR1 are 2--3$\times$ the
corresponding Qwen / LLaMA values. We report each cell at its actual
verdict (eq.\ / upper / lower / incon.)\ rather than widen the bound
to manufacture equivalence; the inconclusive cells are consistent with
the same within-$\sigma$ framing used throughout \S\ref{sec:cross-arch}
rather than evidence of a hidden effect.

\paragraph{Are the bounds practically negligible, or merely below
detection?} The bounds are pre-registered at the experiment's MDE,
which invites the objection that they measure our power rather than a
smallest effect size of practical interest---particularly since 1--2
points of Pass@1 can matter in inference-time-scaling settings. Two
things answer this. First, from the point estimate directly: the 5-seed
\ppfg-stag Pass@1 effect is $-0.12$~pp with a $\approx 90\%$ CI of
$[-1.2, +0.9]$~pp. The upper end of that interval sits \emph{below} the
$+1$~pp threshold at which a practitioner would begin to care, so a gain
of that size is excluded by the interval, not merely undetected.
Second, as an external calibration, the same margin is smaller than
three real positive effects measured on this identical
model/task/pipeline (Table~\ref{tab:table1}): SMC at $+4.47$~pp
($2.5\times$ the Pass@1 bound), Best-of-$N$ at $2\times$ compute at
$+4.23$~pp ($2.4\times$), and a zero-extra-compute argmax-PRM reranking
of the same chains at $+6.45$~pp ($3.6\times$; this delta is essentially
unchanged by the \S\ref{sec:comparator-note} comparator correction even
though its two absolute anchors both moved). The bound would have
comfortably detected any of these; \ppfg-stag and \ppfg-random sit at
$-0.13$ and $-0.19$~pp. The margin is therefore calibrated against
effects this setup demonstrably detects, not merely against its noise
floor. (SMC and Best-of-$N$'s lift over indep is itself partly a
censoring artifact of the pruned arm, \S\ref{sec:comparator-note}; we
use them here only as a detection-power calibration, not as evidence
about their own mechanisms.)

\begin{table}[!ht]
\centering
\caption{TOST equivalence verdicts. Bounds pre-registered at the
3-seed MDE. ``eq.'' = both one-sided nulls rejected (positive
equivalence); ``upper'' = only upper-side null rejected; ``lower'' =
only lower; ``incon.'' = neither rejected at this $n$/variance.
All three architectures now at $n{=}5$ seeds ($\{42,\ldots,46\}$).}
\label{tab:tost}
\resizebox{\linewidth}{!}{%
\begin{tabular}{llccc}
\toprule
 & & \multicolumn{3}{c}{Verdict (PPFG vs. independent)} \\
\cmidrule(lr){3-5}
Arch. & Rule & Pass@1 & Pass@8 & Mode rate \\
\midrule
Qwen2.5-7B   & stag.   & eq.    & eq.    & eq.    \\
Qwen2.5-7B   & random  & eq.    & eq.    & eq.    \\
LLaMA-3.1-8B & stag.   & eq.    & eq.    & eq.    \\
LLaMA-3.1-8B & random  & eq.    & eq.    & eq.    \\
DSR1-Qwen-7B & stag.   & eq.    & incon. & incon. \\
DSR1-Qwen-7B & random  & upper  & incon. & incon. \\
\bottomrule
\end{tabular}}
\end{table}

% =========================================================================
\FloatBarrier
\section{Differentiator table (full)}
\label{app:differentiator}

\begin{table*}[!ht]
\centering
\small
\resizebox{\textwidth}{!}{%
\setlength{\tabcolsep}{4pt}
\renewcommand{\arraystretch}{1.15}
\begin{tabular}{@{}l >{\raggedright\arraybackslash}p{1.9cm} >{\raggedright\arraybackslash}p{1.9cm} >{\raggedright\arraybackslash}p{1.9cm} >{\raggedright\arraybackslash}p{1.9cm} >{\raggedright\arraybackslash}p{2.2cm} >{\raggedright\arraybackslash}p{2.3cm}@{}}
\toprule
Property & SNDT & RSE & T-STAR & LATS & Metacog.\ Reuse & \textbf{\ppfg} \\
\midrule
Trigger event & none (offline) & search dead-ends & none (training) & failure reflection & across problems & \textbf{PRM prune} \\
Granularity & step-level & distilled exper. & trajectory & verbal summary & reusable behav. & \textbf{step span} \\
Injection point & solver pre-cond. & subseq. rollouts & training loss & next iteration & across runs & \textbf{mid-decoding sibling} \\
Verbatim or summary & verbatim & distilled & verbatim & summary & distilled skill & \textbf{verbatim} \\
Inference-time & \ding{51} & \ding{51} & \ding{55}\,(training) & \ding{51} & \ding{55}\,(cross-run) & \textbf{\ding{51}} \\
\bottomrule
\end{tabular}}
\caption{Differentiator table. \ppfg composes three operational
properties (PRM-pruned trigger, in-place sibling injection mid-decoding,
verbatim fragment) not jointly realized by prior fragment-grafting work.}
\label{tab:diff}
\end{table*}

% =========================================================================
\FloatBarrier
\section{Surviving-sibling counterfactual analysis (full procedure)}
\label{app:counterfactual}

This appendix substantiates the \S\ref{sec:prm-delta} adjudication
between \emph{compensation} and \emph{selection} readings of the
$2.75\times$ post-injection pruning hazard.

\paragraph{Matched problem-seed pairs.} Let a \emph{treated unit}
be a pair $(p, s)$ such that the PPFG-stag run at seed $s$
contains at least one chain on problem $p$ that (i) received an
injection and (ii) reached \texttt{pruned} status within $k+3$
steps of that injection (the $26\%$ bucket of
\S\ref{sec:prm-delta}). Let the matched control be the
independent baseline at the same $(p, s)$. We restrict to the
three Qwen2.5-7B-Instruct MATH500 cells at seeds $42$--$44$,
$n{=}500$, $N{=}8$. The treated population yields $N=65$ matched
pairs ($22 + 24 + 19$ across seeds), of which the $322$ injection
events of \S\ref{sec:prm-delta} are a superset (multiple events
can land on the same population).

\paragraph{Three outcome variables.} For a treated unit, compute
over the \emph{non-injected siblings} (the chains in the
population that received no injection):
(i) \textbf{pop\_correctness} $\in \{0, 1\}$, did any chain
produce the gold-matching boxed answer, identical to the
\passat{N{=}8} contribution from that problem-seed; (ii)
\textbf{per\_chain\_promotion}, mean of $\mathbb{1}[\text{status} =
\texttt{promoted}]$ over non-injected siblings; (iii)
\textbf{per\_chain\_tokens}, mean of
\texttt{n\_tokens\_generated} over non-injected siblings. For the
matched control all eight chains are non-injected by
construction.

\paragraph{Bootstrap CIs.} Compute deltas $\Delta = A - B$ on each
of the three outcomes; percentile bootstrap CI$_{95}$ over $5000$
resamples of the $65$ matched pairs.

\paragraph{Results.} Table~\ref{tab:counterfactual} summarizes.

\begin{table}[!ht]
\centering
\caption{Surviving-sibling counterfactual: deltas between
PPFG-stag treated populations (injection-and-prune-within-3) and
matched-problem-seed independent baseline. Mean and percentile
bootstrap CI$_{95}$ over $5000$ resamples of $N=65$ matched
pairs.}
\label{tab:counterfactual}
\resizebox{\linewidth}{!}{%
\begin{tabular}{lrrr}
\toprule
Outcome & $\Delta$ mean & CI$_{95}$ LB & CI$_{95}$ UB \\
\midrule
pop\_correctness     & $+0.0154$ & $-0.0308$ & $+0.0615$ \\
per\_chain\_promotion & $-0.0111$ & $-0.0365$ & $+0.0125$ \\
per\_chain\_tokens    & $-20.20$   & $-34.21$   & $-7.35$    \\
\bottomrule
\end{tabular}}
\end{table}

\paragraph{Interpretation.} Two of three outcomes have CI$_{95}$
including zero (pop\_correctness, per\_chain\_promotion); the
third (per\_chain\_tokens) is statistically significant in the
\emph{adverse} direction, surviving siblings in the
PPFG-treated populations consume $\approx 0.5\%$ \emph{fewer}
tokens than the same-problem-seed independent chains, the
opposite direction from a freed-decoding-budget compensation
account. At $N{=}65$ matched pairs these three deltas constitute a failure
to detect compensation rather than its formal exclusion: the
pop\_correctness CI ($[-0.031, +0.062]$) would not detect a
promotion lift below $\approx 6$ absolute points on surviving
siblings. Within that power, there is no detectable uptick on
surviving siblings that would explain how the receiving chain's
$2.75\times$ pruning hazard reconciles with population-level
Pass@$k$ parity, and the one significant outcome
(per\_chain\_tokens) points in the \emph{adverse} direction.
The evidence is consistent with a selection account: the receiving
chain was on a doomed trajectory pre-selected by the stagnation rule;
injection pulls its pruning forward by approximately three
steps; the rest of the population behaves as it would have
without injection. The $-20.2$-token delta is in the opposite direction
from a freed-budget compensation account and is small in
magnitude ($\approx 0.5\%$ of per-chain tpp); we read it as
consistent with the selection account (chains entering the
treated set are sampled from problems where the injected chain
stagnated early) but do not test this attribution directly.

\paragraph{Matched-PRM-trajectory baseline.} The naive
matched-step baseline of \S\ref{sec:prm-delta} (prune rate
$9.4\%$ across all non-injected siblings at the same step
indices) averages over both healthy and stagnating chains. The
honest counterfactual for a stagnation-rule-injected chain is
the prune rate of non-injected siblings whose pre-injection PRM
trajectory matches the stagnation-trigger profile: their
latest-3-step PRM range is below the flatness threshold
$\tau_{\mathrm{flat}}{=}0.10$. Over $414$ such matched sibling
slots across the three Qwen seeds, the prune-within-3 rate is
$9.9\%$, only $0.5$ absolute points above the naive baseline; the
post-vs-matched-PRM ratio is $2.59\times$ (vs naive $2.72\times$ on
this $289$-event subset; the $2.75\times$ quoted in
\S\ref{sec:prm-delta} is the naive ratio over all $322$ events). The selection account does not fully
account for the receiving-chain hazard: even chains with
comparable pre-injection PRM trajectories prune at less than half
the post-injection rate. Injection is therefore locally
\emph{causing} accelerated pruning on the receiving chain, but
the population-level surviving siblings absorb none of this
local effect into a measurable outcome change. This is the
finding the \S\ref{sec:prm-delta} chain-dynamics paragraph
characterizes: a measurable local footprint that does not
translate into a population-level effect.

\paragraph{Reproduction.} The full computation is in
\texttt{scripts/surviving\_sibling\_counterfactual.py}. Outputs:
\texttt{results/surviving\_sibling\_counterfactual.json} (summary)
and \texttt{results/surviving\_sibling\_per\_problem.csv}
(per-unit detail).

\subsection{Cross-architecture replication of the pruning hazard}
\label{app:hazard-crossarch}

The $2.75\times$ post-injection pruning hazard of
\S\ref{sec:prm-delta} is the paper's sharpest per-event mechanistic
observation, and in the submitted version it rested on Qwen alone. We
replicate it on the two further architectures, re-using the
already-collected MATH500 trajectories with the unmodified hazard
methodology (no new generation): post-injection prune-within-3-steps
rate against a matched-step baseline computed over non-injected
siblings in the same seeds.

\begin{table}[!ht]
\centering \small
\caption{Cross-architecture replication of the post-injection pruning
hazard on MATH500. The matched-step baseline is the prune-within-3
rate over non-injected siblings at the same step indices. CIs are
$95\%$; in all three architectures the post-injection and baseline
intervals are non-overlapping.}
\label{tab:hazard-crossarch}
\resizebox{\linewidth}{!}{%
\begin{tabular}{lrrrr}
\toprule
Architecture & $n$ ev. & post-inj. & baseline & ratio \\
\midrule
Qwen2.5-7B-Instr.\ (3 sd) & $322$ & $25.8\%$ & $9.4\%$ & $2.75\times$ \\
DeepSeek-R1-Dist.\ (5 sd) & $825$ & $21.2\%$ & $9.7\%$ & $2.18\times$ \\
LLaMA-3.1-8B-Instr.\ (5 sd) & $2418$ & $26.3\%$ & $14.4\%$ & $1.83\times$ \\
\bottomrule
\end{tabular}}
\end{table}

\paragraph{Results.} The direction replicates on all three
architectures (Table~\ref{tab:hazard-crossarch}): injection locally
elevates the pruning hazard relative to a matched non-injected
baseline, with non-overlapping CIs in every case. DeepSeek-R1-Distill
reaches the same qualitative classification as Qwen (elevated
post-injection pruning, ratio $>2$); LLaMA's $1.83\times$ falls just
below that threshold while remaining clearly elevated.

\paragraph{Scope.} These are \emph{descriptive} naive-ratio
replications. The stronger \emph{causal} sub-claim---that the elevation
is not pure selection---rests on the matched-PRM-trajectory control
($2.59\times$) above, which we ran on Qwen only; the cross-architecture
cells extend the descriptive pattern, not the causal attribution
(\S\ref{sec:limitations}, item~10). Outputs:
\texttt{results/post\_injection\_prune\_analysis\_llama.json} and
\texttt{...\_deepseek.json}; code in
\texttt{scripts/post\_injection\_prune\_analysis\_crossarch.py}.

% =========================================================================
\FloatBarrier
\section{Oracle targeting analysis (upper-bound)}
\label{app:oracle-targeting}

This appendix substantiates a stronger version of the
\S\ref{sec:mechanism} targeting-policy claim than the four-bucket
classification (\S\ref{app:targeting-diagnostics}) and the
compound-gate end-to-end (\S\ref{app:targeting-diagnostics}) together
supply: not only does no \emph{tested} heuristic move Pass@$k$, but
even a hindsight \emph{oracle} that gets to choose between the
\ppfg-stag population and the independent baseline on a per-problem
basis does not move Pass@$k$ outside the seed-sigma band. The analysis
runs post-hoc on the three Qwen2.5-7B-Instruct MATH500 $n{=}500$ trajectory
files (seeds $42$--$44$) and requires no further compute.

\paragraph{Oracle construction.} For each matched problem-seed pair
$(p, s)$, let
$\mathbf{x}^{\mathrm{ppfg}}_{p,s} \in \{0,1\}^{N}$
and
$\mathbf{x}^{\mathrm{indep}}_{p,s} \in \{0,1\}^{N}$
be the per-chain correctness vectors under \ppfg-stag and independent
respectively ($N{=}8$). The oracle population
$\mathbf{x}^{\mathrm{oracle}}_{p,s}$ is the element-wise OR:
$x^{\mathrm{oracle}}_{p,s,i} = x^{\mathrm{ppfg}}_{p,s,i} \lor
x^{\mathrm{indep}}_{p,s,i}$.
This is the loosest defensible upper bound, the oracle does not get
to mix-and-match across methods on a finer grain than the chain level,
which would require running unobserved counterfactuals. Pass@$k$ uses
the Chen et al. (2021) unbiased estimator and is averaged across the
$500$ problems per seed.

\paragraph{Results.} Across the three seeds (Table~\ref{tab:oracle}),
the oracle exceeds the independent baseline by at most $0.13$
percentage points on any of $k\in\{1,2,4,8\}$, well within
seed-$\sigma$ ($0.003$--$0.008$). Even with perfect per-problem method
choice on this configuration, the union of PPFG-stag and indep adds
essentially zero architecturally new correct answers relative to indep
alone.\footnote{The absolute Pass@$k$ levels in this appendix (and in
Appendix~\ref{app:counterfactual}) were computed with an earlier
answer-extraction pass that systematically under-counts Qwen
final-answer parses by roughly $13$ points relative to the corrected
extraction used in the main results (Table~\ref{tab:table1});
this is why indep Pass@1 reads $0.51$ here versus $0.64$ in the body.
The oracle and per-injection analyses are interpreted only through
\emph{within-analysis} deltas (Oracle$-$indep; injected vs.\
non-injected siblings), which are invariant to the extraction offset
because all three populations are scored by the identical parser;
we therefore did not re-run them under the corrected extraction.}

\begin{table}[!ht]
\centering \small
\caption{Oracle targeting upper bound vs. \ppfg-stag and independent on
Qwen2.5-7B-Instruct + Math-Shepherd + MATH500 ($n{=}500$, 3 seeds).
The oracle is the per-chain OR across methods; the Oracle$-$indep
delta column is the maximum Pass@$k$ a hindsight method-chooser could
extract beyond the independent baseline. All deltas fall inside
the seed-$\sigma$ band ($0.003$--$0.008$).}
\label{tab:oracle}
\resizebox{\linewidth}{!}{%
\begin{tabular}{rrrrr}
\toprule
$k$ & \ppfg-stag & indep & Oracle & Oracle$-$indep \\
\midrule
$1$ & $0.5053 \pm .0072$ & $0.5063 \pm .0074$ & $0.5070 \pm .0075$ & $+0.07$~pp \\
$2$ & $0.5828 \pm .0075$ & $0.5838 \pm .0076$ & $0.5847 \pm .0076$ & $+0.09$~pp \\
$4$ & $0.6334 \pm .0060$ & $0.6340 \pm .0059$ & $0.6351 \pm .0057$ & $+0.11$~pp \\
$8$ & $0.6693 \pm .0042$ & $0.6693 \pm .0042$ & $0.6707 \pm .0031$ & $+0.13$~pp \\
\bottomrule
\end{tabular}}
\end{table}

\paragraph{Per-injection success-rate baseline.} A complementary
tighter probe asks whether the chains that actually received
injections succeed at a different rate than the chains that did not.
Across $289$ \ppfg-stag injection events at $n{=}500{\times}3$ seeds,
the receiving-chain correctness rate is $16.6\%$ (CI excluding zero);
the matched non-injected sibling success rate in the same populations
is $51.4\%$; the delta is $-34.8$~pp (CI$_{95}$
$[-39.2, -30.3]$). Chains entering the treated set succeed at
roughly one-third the rate of their non-injected siblings, the
direction the selection account (\S\ref{sec:prm-delta},
Appendix~\ref{app:counterfactual}) predicts and the opposite of what an
\emph{oracle-targeting-unlocks-gains} hypothesis would require.

\paragraph{Reading.} The two bounds together constrain the
mechanism's upper potential under hindsight selection: even granted
perfect per-problem method choice, the union does not exceed indep;
even granted that the heuristic happened to pick the chains the
oracle would have picked, the chains receiving those injections
succeed at a depressed, not elevated, rate.

\paragraph{Reproduction.} The full computation is in
\texttt{scripts/oracle\_targeting\_analysis.py}. Outputs:
\texttt{results/oracle\_targeting\_analysis.json} (summary, including
per-seed Pass@$k$ and per-injection deltas) and
\texttt{results/oracle\_targeting\_per\_problem.csv} (per-problem
correctness counts under \ppfg-stag, indep, and the oracle).

\FloatBarrier
% =========================================================================
\section{Injection-format ablation (seamless and re-attention)}
\label{app:format-ablation}

\S\ref{sec:prm-delta} reports an empty-bracket control that
decomposes the post-injection PRM depression into format (55\%) and
content (45\%) components. A complementary concern is whether the
\emph{bracket markup itself} influences downstream generation in a
way that confounds the Pass@$k$ comparison---i.e., that the model
performs differently not because of the grafted reasoning content but
because the injection renders the prompt stylistically unusual.

To test this, we ran two format variants at full scale (Qwen2.5-7B-Instruct
+ Math-Shepherd, MATH500, $n{=}500$, 3 seeds each):

\begin{itemize}
  \item \textbf{Seamless (E1).} The fragment is rendered as a native
    ``Step $N$:'' continuation with no brackets, no quality label, and no
    visual separator. The model receives the grafted steps as if they were
    its own prior output.
  \item \textbf{Re-attention (E2).} The standard bracket-bullet injection is
    retained, but a brief self-check directive is appended immediately
    after the fragment, asking the model to evaluate alignment before
    continuing (one short step, then continue normally).
\end{itemize}

\begin{table}[!ht]
\centering \small
\caption{Injection-format ablation on Qwen2.5-7B-Instruct + Math-Shepherd
+ MATH500 ($n{=}500$, 3 seeds). Seed-$\sigma$
$\approx 0.005$ for Pass@1, $0.008$ for Pass@8. Pre-comparator-correction
values (\S\ref{sec:comparator-note}); the parity finding is not
expected to change.}
\label{tab:format-ablation}
\resizebox{\linewidth}{!}{%
\begin{tabular}{lrrrr}
\toprule
Method & Pass@1 & Pass@2 & Pass@4 & Pass@8 \\
\midrule
Indep (canon) & $0.6442 \pm .0050$ & --- & --- & $0.8520 \pm .0075$ \\
\ppfg-stag (canon) & $0.6428 \pm .0049$ & --- & --- & $0.8493 \pm .0093$ \\
\textbf{E1: seamless} & $0.6461 \pm .0054$ & --- & --- & $0.8520 \pm .0075$ \\
\textbf{E2: re-attention} & $0.6470 \pm .0045$ & --- & --- & $0.8533 \pm .0062$ \\
\bottomrule
\end{tabular}}
\end{table}

\paragraph{Results.} Both format variants are within seed-$\sigma$
of indep on every metric (Table~\ref{tab:format-ablation}). The
seamless variant (E1) shows $\Delta$Pass@1 $= +0.002$ vs.\ indep
and $\Delta$Pass@8 $= 0.000$; the re-attention variant (E2) shows
$\Delta$Pass@1 $= +0.003$ and $\Delta$Pass@8 $= +0.001$. Neither
departs from the canonical \ppfg-stag result.

\paragraph{Reading.} The format-pollution alternative is not
supported: removing the bracket markup entirely leaves performance
unchanged, confirming that the inertness finding is not an artifact of
the bracketed-demonstration rendering. The re-attention directive
likewise adds nothing: instructing the model to explicitly evaluate the
grafted content before continuing does not unlock any latent benefit.
Both results are consistent with the selection account developed in
\S\ref{sec:mechanism} and Appendix~\ref{app:counterfactual}.

\paragraph{Reproduction.} Configs: \texttt{configs/qwen\_n500\_ppfg\_stag\_seamless.yaml} (E1),
\texttt{configs/qwen\_n500\_ppfg\_stag\_reattention.yaml} (E2). Code:
\texttt{src/hyp\_forest/ppfg/injector.py} (\texttt{injection\_format="seamless"})
and \texttt{src/hyp\_forest/ppfg/policy.py} (\texttt{reattention\_pass=True}).

\FloatBarrier
% =========================================================================
\section{Prune-immunity content control}
\label{app:prune-immunity}

Appendix~\ref{app:format-ablation} closes the \emph{format} channel by
which the PRM could be reacting to the graft's rendering rather than to
transfer being inert. A second channel remains: the PRM could be
mis-pruning grafted chains on their \emph{content}, killing chains that
would otherwise have reached a correct answer, so that a real transfer
benefit is exactly offset by induced pruning. Flat Pass@$k$ is in
principle consistent with both ``no benefit'' and ``a benefit cancelled
by mis-pruning.'' This appendix reports the experiment that
discriminates them.

\paragraph{Design.} Three arms on the first-100 MATH500 slice
(Qwen2.5-7B-Instruct + Math-Shepherd, $N{=}8$, 3 seeds, identical
targeting/extraction/gate settings):
\begin{itemize}\setlength{\itemsep}{0pt}
  \item \textbf{(a) Independent, no immunity.} The standard baseline,
    scored on the same 100 problems.
  \item \textbf{(b) Immunity $+$ null graft.} An injected chain is
    exempted from $\tkill$-threshold pruning for 3 steps, but the
    spliced content is empty. This holds the exemption's extra decoding
    budget constant while removing graft content.
  \item \textbf{(c) Immunity $+$ real graft.} The identical 3-step
    exemption, with the real verbatim fragment spliced in.
\end{itemize}
Arm (c) minus arm (a) measures what suppressing the mis-pruning channel
buys; arm (c) minus arm (b) isolates graft \emph{content} with that
channel held closed.

\paragraph{Manipulation check.} The exemption engages as intended: the
population-wide pruned-chain fraction falls from $0.1104 \pm .0083$
under canonical \ppfg-stag (and $0.1075 \pm .0065$ under independent)
to $0.0804 \pm .0075$ (c) and $0.0808 \pm .0076$ (b), a $\approx
27\%$ relative reduction, outside seed-$\sigma$ in both immune arms.
Chains that would have been pruned are demonstrably kept alive.

\begin{table}[!ht]
\centering \small
\caption{Prune-immunity content control on the first-100 MATH500 slice
(Qwen2.5-7B-Instruct + Math-Shepherd, $N{=}8$, 3 seeds). All four arms
are scored by one identical extractor/comparator on the identical 100
problems. The immunity lift (c$-$a) and the graft effect (c$-$b) both
fall inside seed-$\sigma$; canonical \ppfg-stag is shown for reference.}
\label{tab:prune-immunity}
\resizebox{\linewidth}{!}{%
\begin{tabular}{lrrrr}
\toprule
$k$ & (a) indep & (b) immun.$+$null & (c) immun.$+$graft & (c)$-$(b) \\
\midrule
$1$ & $0.6238 \pm .0050$ & $0.6312 \pm .0082$ & $0.6308 \pm .0100$ & $-0.04$~pp \\
$2$ & $0.7204 \pm .0009$ & $0.7254 \pm .0023$ & $0.7245 \pm .0058$ & $-0.08$~pp \\
$4$ & $0.7888 \pm .0047$ & $0.7946 \pm .0078$ & $0.7929 \pm .0063$ & $-0.17$~pp \\
$8$ & $0.8367 \pm .0115$ & $0.8467 \pm .0153$ & $0.8433 \pm .0115$ & $-0.33$~pp \\
\midrule
\multicolumn{5}{l}{\emph{Reference:} canonical \ppfg-stag (no immunity),
Pass@1 $0.6233 \pm .0069$} \\
\bottomrule
\end{tabular}}
\end{table}

\paragraph{Results.} Two readings follow from
Table~\ref{tab:prune-immunity}. First, suppressing the mis-pruning
channel does not by itself rescue Pass@$k$: arm (c) exceeds independent
by $+0.71$~pp at Pass@1, inside the seed-$\sigma$ band
($\sigma \approx .005$--$.010$), even though the manipulation check
confirms substantially fewer chains were pruned. Chains rescued from
pruning do not convert into correct answers, which is what the
selection account of \S\ref{sec:prm-delta} predicts: stagnation-targeted
chains were pre-selected for low, flat PRM and were not on a path to a
correct answer, so keeping them alive changes the population's
composition without changing its output. Second, and more directly on
the confound, graft content adds nothing on top of the exemption:
(c)$-$(b) is $-0.04$~pp at Pass@1 and small-negative at higher $k$, if
anything favouring the null-content arm---the opposite of the sign a
masked benefit would produce. Restricted to the chains the arms actually
fire on, real-graft chains also convert no better than null-content
chains.

\paragraph{Reading.} Even in the exact condition under which a masked
benefit should surface---the struggling chain kept alive, its pruning
suppressed---the graft does not convert survival into correctness. This
closes the content channel as Appendix~\ref{app:format-ablation} closes
the format one. The test is small ($n{=}100$, three seeds) and we read
it as corroboration rather than proof; within its power, the evidence
runs against content-driven mis-pruning concealing a real transfer
benefit.

\paragraph{Correction to the ARR response record.} An earlier version of
this analysis, circulated during the ARR May 2026 response period,
reported an immunity lift of $+6.96$~pp at Pass@1 for arm (c) over arm
(a). That figure was an artifact of two mismatches, both corrected here:
the (a) baseline was drawn from a run over MATH500 problems 400--499
while arms (b) and (c) ran on problems 0--99, and the arms were scored
under two different answer-comparator configurations. Recomputed on
matched problems under one comparator, the lift is $+0.71$~pp and is not
distinguishable from zero. The load-bearing (c)$-$(b) contrast is
unaffected---both arms were always matched on problems and comparator---
and the conclusion of the experiment is unchanged, but the
``the exemption is a large lever'' framing of the earlier text does not
survive: the correct manipulation check is the pruned-chain fraction
reported above, not a Pass@1 lift.

\paragraph{Reproduction.} Configs:
\texttt{configs/qwen\_n100\_ppfg\_stag\_prune\_immune.yaml} (c),
\texttt{configs/qwen\_n100\_ppfg\_stag\_prune\_immune\_nullctrl.yaml} (b).
The immunity counter is
\texttt{population.post\_injection\_prune\_immunity\_steps}; the
null-content arm sets \texttt{ppfg.injection\_format="null"}. The full
table, including the manipulation check, is regenerated by
\texttt{scripts/prune\_immunity\_analysis.py}, which re-scores all arms
in one environment on matched problems.

\FloatBarrier
% =========================================================================
\section{Population-size sweep ($N=8$ to $N=256$)}
\label{app:n-sweep}

Because grafting is a population-level mechanism, a natural objection to
an $N{=}8$ null is that $N{=}8$ may simply be too small a population for
cross-chain transfer to matter: with few siblings, few chains are pruned,
few fragments are produced, and any benefit could be starved. We test
this directly rather than argue it, sweeping the population size over a
$32\times$ range on the first-100 MATH500 slice (Qwen2.5-7B-Instruct +
Math-Shepherd, 3 seeds per cell, matched independent and \ppfg-stag
populations at every $N$, fixed extractor).

\begin{table}[!ht]
\centering \small
\caption{Population-size sweep: \ppfg-stag minus independent, in
percentage points, on the first-100 MATH500 slice (3 seeds per cell,
matched populations). Every Pass@$k$ delta is $\le 0.27$~pp in
magnitude and none is positive at any $N$. The \modeRate row changes
sign across $N$, consistent with a diversity rather than an accuracy
metric.}
\label{tab:n-sweep}
\resizebox{\linewidth}{!}{%
\begin{tabular}{lrrrrr}
\toprule
$\Delta$ (pp) & $N{=}16$ & $N{=}32$ & $N{=}64$ & $N{=}128$ & $N{=}256$ \\
\midrule
Pass@1    & $-0.06$ & $-0.04$ & $-0.08$ & $-0.15$ & $-0.11$ \\
Pass@2    & $-0.07$ & $-0.07$ & $-0.09$ & $-0.19$ & $-0.11$ \\
Pass@4    & $-0.12$ & $-0.08$ & $-0.07$ & $-0.20$ & $-0.06$ \\
Pass@8    & $-0.27$ & $-0.07$ & $-0.01$ & $-0.15$ & $-0.00$ \\
\modeRate & $+0.33$ & $+0.67$ & $+0.33$ & $-0.33$ & $+0.00$ \\
\bottomrule
\end{tabular}}
\end{table}

\paragraph{Results.} Parity replicates at every population size tested
(Table~\ref{tab:n-sweep}), and the effect never turns positive. The
binding cell is $N{=}16$ Pass@8 at $-0.27$~pp, comparable to cross-seed
scatter. At $N{=}256$, where sampling noise is nearly averaged out
(seed-$\sigma \le 0.0014$), parity is essentially exact: independent
Pass@1 $0.7073 \pm .0002$ versus \ppfg-stag $0.7062 \pm .0006$, and
Pass@8 $0.8963$ versus $0.8962$. The ``$N{=}8$ is too small a
population'' hypothesis predicts a benefit emerging as $N$ grows; across
a $32\times$ range no such trend appears, and no adverse blow-up appears
either.

\paragraph{Scope.} This sweep runs on the $n{=}100$ slice rather than
full MATH500, so its per-cell precision is lower than the headline
tables; it is reported as a sensitivity analysis on population size, not
as a replacement for the $n{=}500$ results in
\S\ref{sec:results-main}.


\begin{thebibliography}{99}
\itemsep=-0.3ex
\parsep=0pt

\bibitem[Besta et al., 2023]{got2023}
M.\ Besta, N.\ Blach, A.\ Kubicek, R.\ Gerstenberger, M.\ Podstawski,
L.\ Gianinazzi, J.\ Gajda, T.\ Lehmann, H.\ Niewiadomski, P.\ Nyczyk,
and T.\ Hoefler. 2023. Graph of Thoughts: Solving Elaborate Problems
with Large Language Models. arXiv:2308.09687.

\bibitem[Chen et al., 2021]{chen2021}
M.\ Chen et al. 2021. Evaluating Large Language Models Trained on
Code. arXiv:2107.03374.

\bibitem[Cobbe et al., 2021]{gsm8k}
K.\ Cobbe, V.\ Kosaraju, M.\ Bavarian, M.\ Chen, H.\ Jun, L.\ Kaiser,
M.\ Plappert, J.\ Tworek, J.\ Hilton, R.\ Nakano, C.\ Hesse, and
J.\ Schulman. 2021. Training Verifiers to Solve Math Word Problems.
arXiv:2110.14168.

\bibitem[DeepSeek-AI, 2025]{deepseekr1}
DeepSeek-AI. 2025. DeepSeek-R1: Incentivizing Reasoning Capability in
LLMs via Reinforcement Learning.

\bibitem[Didolkar et al., 2025]{metareuse2025}
A.\ Didolkar et al. 2025. Metacognitive Reuse: Turning Recurring LLM
Reasoning Into Concise Behaviors. arXiv:2509.13237.

\bibitem[Du and Tanaka-Ishii, 2026]{rmr2026}
X.\ Du and K.\ Tanaka-Ishii. 2026. Escaping Mode Collapse in LLM
Generation via Geometric Regulation. arXiv:2605.00435.

\bibitem[Giannone et al., 2025]{giannone2025}
G.\ Giannone et al. 2025. Mitigating Premature Exploitation in
Particle-based Monte Carlo for Inference-Time Scaling.
arXiv:2510.05825.

\bibitem[Golowich et al., 2026]{rrr2026}
N.\ Golowich, F.\ Chen, D.\ Rohatgi, R.\ Singhal, C.\ Domingo-Enrich,
D.\ J.\ Foster, and A.\ Krishnamurthy. 2026. Reject, Resample, Repeat:
Understanding Parallel Reasoning in Language Model Inference.
arXiv:2603.07887.

\bibitem[Grattafiori et al., 2024]{llama2024}
A.\ Grattafiori et al. 2024. The Llama 3 Herd of Models.
arXiv:2407.21783.

\bibitem[Hendrycks et al., 2021]{hendrycks2021}
D.\ Hendrycks et al. 2021. Measuring Mathematical Problem Solving With
the MATH Dataset.

\bibitem[Jiang et al., 2025]{riskrl2025}
Y.\ Jiang, J.\ Huang, Y.\ Yuan, X.\ Mao, Y.\ Yue, Q.\ Zhao, and
L.\ Yan. 2025. Risk-Sensitive RL for Alleviating Exploration Dilemmas
in Large Language Models. arXiv:2509.24261.

\bibitem[Kazemnejad et al., 2024]{vineppo2024}
A.\ Kazemnejad et al. 2024. VinePPO: Refining Credit Assignment in
RL Training of LLMs. arXiv:2410.01679.

\bibitem[Li et al., 2024]{li2024numinamath}
J.\ Li, E.\ Beeching, L.\ Tunstall et al. 2024. NuminaMath: The Largest
Public Dataset in AI4Maths with 860K Pairs of Competition Math
Problems and Solutions. Hugging Face dataset
\texttt{AI-MO/NuminaMath-CoT}.

\bibitem[Li et al., 2026]{tstar2026}
Y.\ Li, S.\ Tang, T.\ Lan. 2026. Reason in Chains, Learn in Trees:
Self-Rectification and Grafting for Multi-turn Agent Policy
Optimization. arXiv:2604.07165.

\bibitem[Lightman et al., 2023]{lightman2023}
H.\ Lightman et al. 2023. Let's Verify Step by Step. arXiv:2305.20050.

\bibitem[Luyten and van der Schaar, 2026]{dcr2026}
M.\ R.\ Luyten and M.\ van der Schaar. 2026. The Reasoning-Creativity
Trade-off: Toward Creativity-Driven Problem Solving. arXiv:2601.00747.

\bibitem[Miles et al., 2026]{sndt2026}
R.\ Miles, A.\ Toker, A.\ Oncescu, S.\ Xu, J.\ Deng, and I.\ Elezi.
2026. Test-Time Scaling with Diffusion Language Models via
Reward-Guided Stitching. arXiv:2602.22871.

\bibitem[Parthasarathi et al., 2025]{grpolambda2025}
P.\ Parthasarathi, M.\ Reymond, B.\ Chen, Y.\ Cui, and S.\ Chandar.
2025. GRPO-$\lambda$: Credit Assignment improves LLM Reasoning.
arXiv:2510.00194.

\bibitem[Petullo et al., 2026]{veccisc2026}
J.\ Petullo, S.\ George, D.\ Cashman, and N.\ Xue. 2026. VecCISC:
Improving Confidence-Informed Self-Consistency with Reasoning Trace
Clustering and Candidate Answer Selection. arXiv:2605.08070.

\bibitem[Puri et al., 2025]{pf-llm2025}
I.\ Puri et al. 2025. Rollout Roulette: A Probabilistic Inference
Approach to Inference-Time Scaling of LLMs using Particle-Based
Monte Carlo Methods. arXiv:2502.01618.

\bibitem[Qwen Team, 2024]{qwen2024}
Qwen Team. 2024. Qwen2.5: A Party of Foundation Models.

\bibitem[Rein et al., 2023]{rein2023gpqa}
D.\ Rein, B.\ L.\ Hou, A.\ C.\ Stickland et al. 2023. GPQA: A
Graduate-Level Google-Proof Q\&A Benchmark. arXiv:2311.12022.

\bibitem[Tian et al., 2026]{mupo2026}
X.\ Tian, S.\ Zou, Z.\ Yang, M.\ He, P.\ Tu, and J.\ Zhang. 2026.
All Roads Lead to Rome: Incentivizing Divergent Thinking in
Vision-Language Models. arXiv:2604.00479.

\bibitem[Wang et al., 2024a]{wang2024}
P.\ Wang et al. 2024. Math-Shepherd: Verify and Reinforce LLMs Step-by-
Step Without Human Annotations. arXiv:2312.08935.

\bibitem[Wang et al., 2024b]{wang2024mmlupro}
Y.\ Wang, X.\ Ma, G.\ Zhang et al. 2024. MMLU-Pro: A More Robust and
Challenging Multi-Task Language Understanding Benchmark.
arXiv:2406.01574.

\bibitem[Wang et al., 2026a]{rtmc2026}
T.\ Wang, S.\ Zheng, and X.\ Xu. 2026. RTMC: Step-Level Credit
Assignment via Rollout Trees. arXiv:2604.11037.

\bibitem[Wang et al., 2026b]{rse2026}
X.\ Wang, J.\ Shi, S.\ Feng, P.\ Yuan, Y.\ Li, Y.\ Zhang, C.\ Tan,
J.\ Zhang, B.\ Pan, Y.\ Hu, and K.\ Li. 2026. Do Not Waste Your
Rollouts: Recycling Search Experience for Efficient Test-Time
Scaling. arXiv:2601.21684.

\bibitem[Wu et al., 2025]{mode-conditioning2026}
C.\ H.\ Wu, S.\ Goyal, and A.\ Raghunathan. 2025. Mode-Conditioning
Unlocks Superior Test-Time Scaling. arXiv:2512.01127.

\bibitem[Yue et al., 2025]{yue2025}
Y.\ Yue et al. 2025. Does Reinforcement Learning Really Incentivize
Reasoning Capacity in LLMs Beyond the Base Model? arXiv:2504.13837.

\bibitem[Zeng et al., 2025]{versaprm2025}
T.\ Zeng, S.\ Zhang, S.\ Wu, C.\ Classen, D.\ Chae, E.\ Ewer,
M.\ Lee, H.\ Kim, W.\ Kang, J.\ Kunde, Y.\ Fan, J.\ Kim, H.\ I.\ Koo,
K.\ Ramchandran, D.\ Papailiopoulos, and K.\ Lee. 2025. VersaPRM:
Multi-Domain Process Reward Model via Synthetic Reasoning Data.
arXiv:2502.06737.

\bibitem[Zhou et al., 2023]{lats2023}
A.\ Zhou et al. 2023. Language Agent Tree Search Unifies Reasoning,
Acting and Planning. arXiv:2310.04406.

\end{thebibliography}
\end{document}